\documentclass{article}
\usepackage[final]{colm2026_conference}

\usepackage{microtype}
\usepackage{hyperref}
\usepackage{url}
\usepackage{booktabs}
\usepackage{array}   
\usepackage{graphicx}
\usepackage{subcaption}
\usepackage{multirow}
\usepackage{arydshln}
\usepackage{amsmath}
\usepackage{amssymb}
\usepackage{xcolor}
\usepackage{pifont}
\usepackage{wrapfig}
\usepackage[capitalize,noabbrev]{cleveref}
\usepackage{lineno}

\definecolor{darkblue}{rgb}{0, 0, 0.5}
\hypersetup{colorlinks=true, citecolor=darkblue, linkcolor=darkblue, urlcolor=darkblue}

\newcommand{\coderepo}{https://github.com/jiaweixu98/SEER}

\title{\textsc{Seer}: Long-Context Reasoning via Selective \mbox{Visual-Text Compression}}

\author{
\textbf{Jiawei Xu$^{1,*}$ \quad Zhilin Zhai$^{2,*}$ \quad Jinrui Fang$^{1}$ \quad Ruohan Xu$^{3}$ \quad Mingfei Lu$^{4}$} \\
\textbf{Yi Zhang$^{4}$ \quad Guanchu Wang$^{5,\dagger}$ \quad Tianlong Chen$^{2,\dagger}$ \quad Ying Ding$^{1,\dagger}$} \\[2pt]
$^{1}$School of Information, The University of Texas at Austin \\
$^{2}$Department of Computer Science, The University of North Carolina at Chapel Hill \\
$^{3}$Department of Engineering, University of Cambridge \\
$^{4}$Faculty of Engineering and Information Technology, University of Technology Sydney \\
$^{5}$Department of Computer Science, The University of North Carolina at Charlotte \\[2pt]
$^{*}$Equal contribution. \quad $^{\dagger}$Equal senior authorship. \\[2pt]
\small\texttt{jiaweixu@utexas.edu} \quad \texttt{zzhilin@unc.edu} \quad \texttt{jinrui@utexas.edu} \\
\small\texttt{rx232@cam.ac.uk} \quad \texttt{mingfei.lu@student.uts.edu.au} \quad \texttt{Yi.Zhang@uts.edu.au} \\
\small\texttt{guanchu.wang@charlotte.edu} \quad \texttt{tianlong@cs.unc.edu} \quad \texttt{ying.ding@ischool.utexas.edu}
}

\begin{document}

\ifcolmsubmission
\linenumbers
\fi

\maketitle

\begin{abstract}
Long-context reasoning remains computationally expensive for large language models due to the quadratic complexity of attention over text tokens.
Visual-text compression offers a promising alternative by rendering text into images and processing them with vision-language models, often reducing token usage.
However, existing approaches apply uniform compression regardless of query relevance, potentially sacrificing precision where detailed extraction is required.
We present \textsc{Seer}, a framework that learns to \emph{select} query-relevant images through visual scanning and \emph{retrieve} textual content only where needed, combining the efficiency of visual compression with the precision of text-based reasoning.
Through supervised fine-tuning on tool-interaction trajectories, \textsc{Seer} learns adaptive tool invocation for selection and retrieval.
Experiments on long-context benchmarks show that \textsc{Seer} improves extraction precision through selective text retrieval while retaining average prompt-token savings relative to full-text baselines. On LongBench, \textsc{Seer} achieves 51.11\% average accuracy, outperforming the visual-text baseline Glyph-9B by 2.33 points and Qwen3-8B by 3.49 points.
Code can be accessed at \href{\coderepo}{\texttt{\coderepo}}.

\end{abstract}

\section{Introduction}
Long-context reasoning~\citep{bai2024longbench, sui2025stop} is increasingly important for applications such as multi-turn dialogue~\citep{maharana2024evaluating} and long-horizon tasks~\citep{deng2025swe, merrill2026terminalbench}.
However, processing long sequences with large language models (LLMs) remains computationally expensive due to the quadratic complexity of self-attention~\citep{vaswani2017attention}.
This challenge has motivated various approaches, including extended positional encodings such as YaRN~\citep{peng2024yarn,wu2024extending} and sparse or linear attention~\citep{yang2024gated}.

Recently, visual-text compression has emerged as a promising paradigm for efficient long-context processing~\citep{cheng2026glyph,xing2025visioncentric,wei2025deepseek}.
By rendering textual content into images and processing them with vision-language models (VLMs), this approach can provide substantial token compression (number of textual tokens to vision tokens, typically 3--4$\times$ in prior work) while maintaining competitive accuracy~\citep{cheng2026glyph,luo2026autol2s}.
The key insight is that vision encoders can represent dense textual information more compactly than tokenized text, enabling faster prefilling.

\begin{wrapfigure}[32]{r}{0.50\textwidth}
  \centering
  \includegraphics[width=\linewidth]{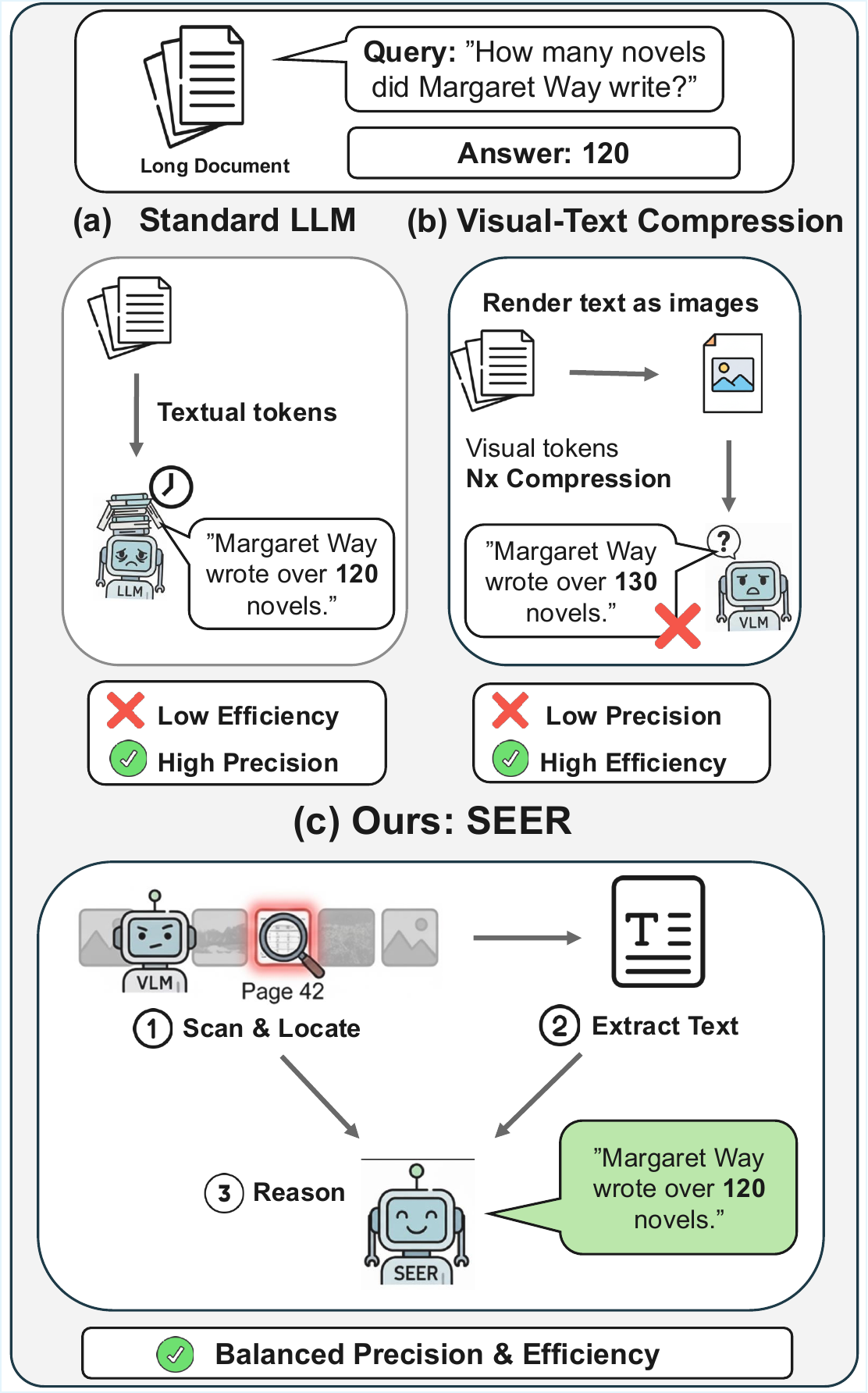}
  \caption{Long-context QA trade-off among LLMs, visual-text compression, and \textsc{Seer}.}
  \label{fig:teaser}
\end{wrapfigure}
Despite these advantages, existing visual-text compression methods apply uniform compression across the entire context, regardless of query relevance~\citep{cheng2026glyph}.
This uniform treatment presents a practical trade-off: aggressive compression improves efficiency but may lose fine-grained details necessary for precise reasoning; conservative compression preserves details but sacrifices efficiency gains.
We observe that this trade-off can often be alleviated when query relevance~\citep{liu2024lost,jiang2024longllmlingua} is considered: many queries require detailed information from only a small subset of the context, while the remaining content serves primarily for global understanding (see ablation in \cref{sec:ablation-selection}).

We propose \textsc{Seer}, a framework that learns selective visual-text compression by treating visual selection and text retrieval as tools within a unified reasoning process (\cref{fig:teaser}).
Our key insight is that visual representations enable rapid global information access, analogous to how humans skim a document to identify relevant sections, while textual representations support precise reasoning.
\textsc{Seer} leverages this complementarity through a three-stage process: (1) \emph{select} query-relevant images through visual scanning, (2) \emph{retrieve} the corresponding source text (available from the rendering process), and (3) \emph{reason} over both visual context and retrieved text to generate the answer.
By learning which images require precise textual information, \textsc{Seer} aims to balance the efficiency of visual compression with precision where needed.
Our contributions are summarized as follows:

\begin{itemize}
  \item We propose \textsc{Seer}, a framework that combines visual-text compression with selective tool invocation for long-context reasoning, learning to retrieve text only from selected images that are most relevant to inference.
  \item We propose a three-stage inference pipeline with a training strategy that jointly optimizes selection and reasoning capabilities.

  \item On LongBench, \textsc{Seer} achieves 51.11\% average accuracy, improving over Glyph-9B by 2.33 points through selective text retrieval, with large gains on extraction-intensive tasks (+8.75 on Qasper, +19.17 on MultiFieldQA-Zh). Under our reported setup, \textsc{Seer} is 3.49 points above Qwen3-8B.
\end{itemize}
\section{Related Work}

\paragraph{Visual-Text Compression.}
Recent work has explored compressing textual inputs by rendering long texts into images and processing them with vision-language models (VLMs), offering a promising direction for improving long-context reasoning efficiency.
Glyph~\citep{cheng2026glyph} achieves 3 to 4$\times$ token compression while maintaining accuracy comparable to leading LLMs on various long-context benchmarks, reporting approximately 4$\times$ faster prefilling and decoding, and 2$\times$ faster supervised fine-tuning.
Vision-centric Token Compression (Vist)~\citep{xing2025visioncentric} introduces a slow-fast compression framework inspired by human reading behavior: the fast path renders distant tokens into images and employs a lightweight vision encoder to skim low-salience context, while the slow path processes proximal tokens through the LLM for fine-grained reasoning.
DeepSeek-OCR~\citep{wei2025deepseek} investigates the feasibility of compressing long contexts via optical 2D mapping, demonstrating that when the compression ratio remains below 10$\times$ (i.e., text tokens are within 10 times the number of vision tokens), the model can achieve 97\% OCR precision.
Across these methods the granularity of compression is fixed in advance, either as a global ratio or by token position, and never as a function of what a particular query needs. That is the gap \textsc{Seer} addresses.

\paragraph{Tool-integrated Reasoning.}
Chain-of-Thought (CoT) prompting~\citep{wei2022chain} established a foundational paradigm for complex problem solving through explicit decomposition into stepwise reasoning traces~\citep{guo2025deepseek, jaech2024openai}, where each rationale incrementally builds upon previous steps.
Building on this foundation, tool-integrated reasoning architectures enhance model capabilities through external tool integration.
Prompting-based methods such as IRCoT~\citep{trivedi2023interleaving} and ReAct~\citep{yao2022react} initiate tool-reasoning loops via structured templates, with ReAct establishing an influential paradigm through iterative Thought-Action-Observation cycles that sequentially sample reasoning thoughts, tool invocations, and environmental responses~\citep{sumers2024cognitive}.
However, these approaches face scalability challenges in open-domain environments, where the combinatorial growth in action space can make tool selection difficult. More recent work leverages supervised fine-tuning and reinforcement learning to learn flexible tool invocation strategies, enabling dynamic tool selection within reasoning processes~\citep{qian2025toolrl,luo2026demystifying}.
\textsc{Seer} instantiates this paradigm with visual selection and text retrieval as its two tools.

\section{Problem Setup}
\label{sec:problem-setup}

We study long-context question answering with visual-text compression.
Given a textual context $\mathcal{C}$ and a query $q$, the goal is to produce an answer $y$.

\paragraph{Visual-Text Compression.}
Rather than processing all tokens through LLMs, visual-text compression first renders the textual context $\mathcal{C}$ into images:
\begin{equation}
    \mathcal{I} = \operatorname{Render}(\mathcal{C}) = (I_1, I_2, \ldots, I_m),
\end{equation}
where each image $I_j$ contains a rendered segment of the textual context.
A VLM $f$ then processes these images along with the query:
\begin{equation}
    y = f(\mathcal{I}, q).
\end{equation}
This corresponds to the uniform-compression baseline. In contrast, our setting selectively retrieves source text from query-relevant images before generating the final answer.

Let $\tau_{\text{text}}$ denote the number of text tokens in the textual context $\mathcal{C}$ and $\tau_{\text{vision}}$ the total number of vision tokens produced by the vision encoder over $\mathcal{I}$.
This approach can yield substantial compression because vision encoders typically represent each image with far fewer tokens than the original text, with prior work reporting compression ratios $\tau_{\text{text}} / \tau_{\text{vision}}$ of $3$--$4\times$ while maintaining competitive performance~\citep{cheng2026glyph,yuan2024kv}.

\paragraph{Limitations of Uniform Compression.}
However, this paradigm processes all visual content uniformly, regardless of its relevance to the query.
For tasks requiring precise information extraction from specific images, the compressed visual representation may lack sufficient granularity for accurate reasoning.
Our objective is to answer $q$ accurately while selectively extracting text only from query-relevant images, thereby avoiding full-text decoding.

\section{Method}
\label{sec:method}

Following \cref{sec:problem-setup}, we start from a textual context $\mathcal{C}$ and query $q$, and render $\mathcal{C}$ into images $\mathcal{I}=\operatorname{Render}(\mathcal{C})$.
Building on the intuition that visual and textual representations offer complementary strengths, we introduce \textsc{Seer}, an end-to-end framework that \emph{selects} key images, \emph{retrieves} relevant textual context from the selection, and \emph{reasons} over the visual-text representation.

\subsection{Select-Retrieve-Reason Framework}
\label{sec:framework}
\textsc{Seer} mitigates accuracy loss from uniform compression through a unified tool-interaction trajectory of \emph{Selection}, \emph{Retrieval}, and \emph{Reasoning}. For simplicity, we use $f$ to denote the same autoregressive model across these generation stages.

\paragraph{Stage 1: Visual Selection.} Given the image sequence $\mathcal{I} = (I_1, \ldots, I_m)$ and query $q$, the model $f$ predicts a set of relevant indices $S \subseteq \{1, \ldots, m\}$, corresponding to selected images $\boldsymbol{y}_S = \{I_j \mid j \in S\}$:
\begin{equation}
    S = f(\mathcal{I}, q).
\end{equation}
This leverages the global accessibility of visual representations for rapid selection of query-relevant regions.
In implementation, predicted indices are canonicalized before retrieval: we remove duplicates and keep only valid indices in $[1,m]$.
Out-of-range indices are discarded (rather than corrected).

\paragraph{Stage 2: Text Retrieval.}
For the selected images $\boldsymbol{y}_S$, the system retrieves the corresponding source text:
\begin{equation}
    \boldsymbol{y}_R = \operatorname{Retrieve}(\boldsymbol{y}_S).
\end{equation}
Since the source text is available from the rendering process, this is a deterministic lookup rather than model inference. This leverages textual representations for fine-grained local reasoning.

\paragraph{Why Retrieval Instead of OCR?}
We adopt deterministic text retrieval rather than training an OCR component for three reasons:
(1) \textbf{Task focus}: Our core contribution is learning \emph{when} precise text is needed, not \emph{how} to extract it.
(2) \textbf{Practical relevance}: In many digital-document settings (PDFs, web pages, rendered code), source text is available; OCR is mainly needed for scanned documents.
(3) \textbf{Modularity}: This design allows drop-in replacement with OCR systems when source text is unavailable, without retraining the selection model.

\paragraph{Stage 3: Reasoning.}
The model $f$ generates the response by reasoning over the combined context of the selected images $\boldsymbol{y}_S$ and retrieved text $\boldsymbol{y}_R$:
\begin{equation}
    \boldsymbol{y}_{\text{CoT}} = f(\mathcal{I}, q, \boldsymbol{y}_S, \boldsymbol{y}_R).
\end{equation}
In this stage, the model $f$ retains access to all visual images $\mathcal{I}$ (for global context) along with the retrieved text $\boldsymbol{y}_R$ (for precise local information from query-relevant regions).
Conditioned on this reasoning trace, the model then generates the final answer:
\begin{equation}
    y = f(\mathcal{I}, q, \boldsymbol{y}_S, \boldsymbol{y}_R, \boldsymbol{y}_{\text{CoT}}).
\end{equation}

\paragraph{Tool-Based Formulation.}
We implement this as a tool call: the model $f$ outputs a structured invocation specifying which images to select, receives the text as an observation, and continues reasoning.
This follows the ReAct-style Thought-Action-Observation paradigm~\citep{yao2022react}.
See Appendix~\ref{app:special-tokens} for special token definitions and Appendix~\ref{app:data-format} for complete trajectory examples.

\subsection{Training Strategy}
\label{sec:training}

We train \textsc{Seer} with supervised fine-tuning (SFT) on complete Select-Retrieve-Reason tool-interaction trajectories.
We initialize from Glyph-9B~\citep{cheng2026glyph}.
Full implementation details, including model architecture and hyperparameters, are provided in Appendix~\ref{app:implementation} and \ref{app:hyperparameters}.

\paragraph{Training Data Construction.}
We construct a dataset $\mathcal{D} = \{(\mathcal{I}^{(i)}, q^{(i)}, \mathcal{I}^{*(i)}, r^{(i)}, y^{(i)})\}_{i=1}^{N}$, where $N = |\mathcal{D}|$ is the number of training samples, $\mathcal{I}^{(i)}$ is the context image sequence, $q^{(i)}$ is the query, $\mathcal{I}^{*(i)}$ denotes the ground-truth relevant images, $r^{(i)}$ is the reference chain-of-thought reasoning text, and $y^{(i)}$ is the answer.
To obtain relevance annotations $\mathcal{I}^{*(i)}$, we use Qwen3.5-9B to assess each image independently: given the query, the reference answer, and an image's textual content, it determines whether the image is necessary to derive the answer.
The reference answer is used only during offline data annotation and quality filtering, and is never provided at inference or evaluation time.
We then use Qwen3-32B, given the query, textual context, and selected-image context, to generate the reasoning chain (CoT) and final answer.
We retain only samples where Qwen3-32B answers correctly.
Therefore, each training trajectory consists of (1) the rendered image sequences, (2) the query and instruction, (3) a tool call specifying which images to select (e.g., \texttt{\{"name": "select", "indices": [1, 3]\}}), (4) the retrieved text as an observation, and (5) chain-of-thought reasoning followed by the final answer.

Specifically, we construct training trajectories from the \emph{train splits} of 2WikiMQA, HotpotQA, GovReport, LCC-Python, MuSiQue, Qasper, and TriviaQA to avoid data leakage while maintaining dataset diversity (see Appendix~\ref{app:data-construction} for details).

\paragraph{Supervised Fine-Tuning (SFT).}
Given the training dataset $\mathcal{D}$ constructed above, we train \textsc{Seer} on complete trajectories (select $\rightarrow$ observation $\rightarrow$ reasoning $\rightarrow$ answer).
We use a per-token weight mask to supervise only model-generated content. Let $x_{1:T}$ denote a tokenized trajectory and $w_t \in \mathbb{R}_{\ge 0}$ denote the weight for token $x_{t+1}$: $w_t=0$ for non-generated observation tokens, and $w_t \ge 1$ for generated tokens (with optional upweighting for tool-call and format-related tokens).
The SFT objective is:
\begin{equation}
    \mathcal{L}_{\text{SFT}} = \mathbb{E}_{(x,w)\sim\mathcal{D}}\!\left[-\frac{\sum_{t=1}^{T-1} w_t \log p_\theta(x_{t+1}\mid x_{\le t})}{\sum_{t=1}^{T-1} w_t}\right],
\end{equation}
where $p_\theta$ denotes the autoregressive next-token distribution.
In implementation, we apply this weighted normalization within each micro-batch.

\section{Experiments}

We evaluate \textsc{Seer} on long-context QA benchmarks, comparing against both text-based long-context LLMs and visual-text compression methods.

\paragraph{Datasets.}
We evaluate on LongBench~\citep{bai2024longbench}, a bilingual benchmark with 21 tasks spanning 6 categories: single-document QA, multi-document QA, summarization, few-shot learning, synthetic tasks, and code completion. LongBench contains 4,750 evaluation samples in total. The average context length is 6,711 words for English and 13,386 characters for Chinese.

\paragraph{Baselines.}
We report three groups of baselines, prioritizing open-weight models with comparable parameter scales (roughly 7B--9B) and similar release periods.
Our primary controlled comparison is against Glyph-9B~\citep{cheng2026glyph} (same base architecture family and visual-compression pipeline), while comparisons to external text-based LLMs are reported for practical context.
\textbf{(1) Text-based LLMs with extended context windows:} LLaMA-3.1-8B-Instruct~\citep{grattafiori2024llama}, Qwen2.5-7B-Instruct-1M~\citep{yang2025qwen25_1m}, Qwen3-8B~\citep{yang2025qwen3}, and GLM-4-9B-Chat-1M~\citep{glm2024chatglm}. These models process full text directly.
\textbf{(2) Visual-text compression:} Glyph-9B, which is also our initialization model. Glyph renders text as images and performs pure visual reasoning without selective text retrieval.
\textbf{(3) Proprietary reference point:} we additionally list GPT-4.1 in gray. It is a large closed-weight model well beyond our parameter budget, so we report it only to situate the 7B--9B results on a common scale and exclude it from the best/second-best comparison.
Because \textsc{Seer} is task-specifically fine-tuned on our constructed trajectories while the text-based baselines are evaluated in direct inference mode, cross-family comparisons are not fully iso-training.

\paragraph{Implementation Details.}
We train \textsc{Seer} on 2$\times$ H100 80GB GPUs using the Verl framework.
All \textsc{Seer} checkpoints reported in this paper are initialized from Glyph-9B and then fine-tuned with our Select-Retrieve-Reason trajectories.
In the current setting, SFT runs for 6{,}000 optimizer steps with learning rate $3 \times 10^{-6}$, global batch size 2, and maximum sequence length 16,384.
Following \cref{sec:training}, we use a weighted token-level SFT objective with non-generated observation tokens masked out.
We use fully sharded data parallelism with gradient checkpointing.

\paragraph{Evaluation Metrics.}
We follow the official LongBench evaluation protocol and script~\citep{bai2024longbench} with dataset-specific metric mapping.
Specifically, QA uses F1 variants, summarization uses ROUGE variants, classification uses \texttt{classification\_score}, retrieval uses \texttt{retrieval\_score}/\texttt{retrieval\_zh\_score}, \texttt{passage\_count} uses \texttt{count\_score}, and code tasks (\texttt{lcc}, \texttt{repobench-p}) use \texttt{code\_sim\_score}.
Following the official LongBench convention, the final \textbf{Avg}/\textbf{Overall} score is computed as a \emph{macro-average over major task categories}, rather than a micro-average over all individual datasets.
Concretely, if $C$ is the set of major categories and $D_c$ is the set of datasets in category $c$, we compute
\[
\mathrm{Avg}=\frac{1}{|C|}\sum_{c\in C}\left(\frac{1}{|D_c|}\sum_{d\in D_c}\mathrm{Score}_d\right).
\]
This design gives equal weight to each core capability category even when categories contain different numbers of datasets.

\section{Results}

\subsection{Main Results}

\begin{table*}[htbp]
    \centering
    \resizebox{1\linewidth}{!}{%
    \begin{tabular}{@{}l cccccccccccc@{}}
    \toprule
    \multirow{2}{*}{\textbf{Model}} 
    & \multicolumn{2}{c}{\textbf{Single-Doc QA}} 
    & \multicolumn{2}{c}{\textbf{Multi-Doc QA}} 
    & \multicolumn{2}{c}{\textbf{Summarization}} 
    & \multicolumn{2}{c}{\textbf{Few-shot}} 
    & \multicolumn{3}{c}{\textbf{Synthetic}}
    & \multirow{2}{*}{\textbf{Avg}} \\
    \cmidrule(lr){2-3} \cmidrule(lr){4-5} \cmidrule(lr){6-7} \cmidrule(lr){8-9} \cmidrule(lr){10-12}
    & \textbf{QP} & \textbf{NQA} & \textbf{HQA} & \textbf{2QA} & \textbf{QSUM} & \textbf{GovRep} & \textbf{TREC} & \textbf{TriQA} & \textbf{PR Zh} & \textbf{PR En} & \textbf{Pa C} & \\
    \midrule
    \multicolumn{13}{l}{\textit{\small Text-based Inference}} \\[-4pt]
    \midrule
    \textcolor{gray}{GPT-4.1} & \textcolor{gray}{51.60} & \textcolor{gray}{35.73} & \textcolor{gray}{69.10} & \textcolor{gray}{74.15} & \textcolor{gray}{23.50} & \textcolor{gray}{33.36} & \textcolor{gray}{77.00} & \textcolor{gray}{93.36} & \textcolor{gray}{100.00} & \textcolor{gray}{100.00} & \textcolor{gray}{26.50} & \textcolor{gray}{56.56} \\
    \hdashline\addlinespace[2pt]
    LLaMA-3.1-8B-Instruct & 44.56 & 26.34 & 56.88 & 46.67 & \textbf{23.28} & \textbf{32.36} & 19.25 & \underline{89.12} & 62.20 & \underline{99.50} & 7.13 & 40.52 \\
    Qwen2.5-7B-Instruct-1M & \underline{45.29} & 25.61 & \underline{60.70} & 40.51 & \underline{22.95} & \underline{29.97} & 59.37 & 86.93 & \underline{98.50} & \textbf{100.00} & 3.50 & 42.41 \\
    Qwen3-8B & 43.86 & \underline{27.08} & \textbf{65.60} & \textbf{73.62} & 19.68 & 26.97 & 72.00 & 89.11 & \textbf{100.00} & 96.88 & \underline{11.57} & 47.62 \\
    GLM-4-9B-Chat-1M & 43.70 & \textbf{27.14} & 57.67 & 51.64 & 22.81 & 27.73 & 61.52 & 88.91 & \textbf{100.00} & \underline{99.50} & 0.50 & 44.44 \\
    \midrule
    \multicolumn{13}{l}{\textit{\small Visual-Text Compression}} \\[-4pt]
    \midrule
    Glyph-9B & 38.99 & 21.83 & 56.42 & \underline{71.62} & 19.03 & 25.18 & \textbf{79.50} & 88.04 & 91.00 & 90.50 & \textbf{30.00} & \underline{48.78} \\
    \textbf{Seer (Ours)} & \textbf{47.74} & 26.05 & 60.22 & 68.65 & 20.26 & 27.30 & \underline{78.00} & \textbf{89.76} & 97.00 & 88.25 & \textbf{30.00} & \textbf{51.11} \\
    \bottomrule
    \end{tabular}%
    }
    \caption{Main LongBench results (\%). \textbf{Avg} is the official category-level macro-average computed over all 21 tasks (the remaining task columns, including the Code category, are reported in \Cref{tab:longbench-subtasks}). Glyph-9B uses pure visual reasoning. \textsc{Seer} augments visual reasoning with selective text retrieval and shows improved accuracy on several extraction-heavy tasks. The row in \textcolor{gray}{gray} (GPT-4.1) is a large proprietary reference point included only to situate the open-weight results on a common scale; it is excluded from the ranking. Among the remaining models, \textbf{bold} marks the best and \underline{underline} the second-best value per column.}
    \label{tab:longbench-main}
\end{table*}

\Cref{tab:longbench-main} presents the main LongBench results, while the remaining task columns are reported in \Cref{tab:longbench-subtasks} (Appendix~\ref{app:longbench-subtasks}).
\textsc{Seer} achieves an overall average of 51.11\%, improving over the base model Glyph-9B (48.78\%) by 2.33 points.
Under our reported setup, it is also 3.49 points above Qwen3-8B (47.62\%).

\textbf{Selective retrieval is associated with gains on extraction-heavy tasks.} Notable gains appear on tasks requiring precise information extraction.
On single-document QA, \textsc{Seer} improves over Glyph-9B by +8.75 on Qasper (38.99$\to$47.74) and +4.22 on NarrativeQA (21.83$\to$26.05); on multi-document QA, HotpotQA improves by +3.80 (56.42$\to$60.22).
Single-document QA subtasks in \Cref{tab:longbench-subtasks} show particularly large gains (+19.17 on QA~Zh and +10.76 on QA~En), suggesting that selective text retrieval can be especially useful when densely rendered Chinese text introduces greater visual ambiguity.
These extraction-intensive tasks share a common pattern: the answer hinges on precise details (names, numbers, dates) that benefit from textual over visual processing.

\subsection{Token Reduction}
\label{sec:token-reduction}

\begin{table*}[t]
    \centering
    \setlength{\tabcolsep}{3pt}
    \resizebox{\textwidth}{!}{%
    \begin{tabular}{lcccccc}
    \toprule
    \multirow{3}{*}{\textbf{Model}}
    & \multicolumn{4}{c}{\textbf{Prompt Tokens}}
    & \multicolumn{2}{c}{\textbf{Output Efficiency}} \\
    \cmidrule(lr){2-5}\cmidrule(lr){6-7}
    & \textbf{Textual Prompt}
    & \multirow{2}{*}{\textbf{Visual Prompt}}
    & \multirow{2}{*}{\textbf{Retrieved Text}}
    & \multirow{2}{*}{\textbf{Total Prompt}}
    & \textbf{Completion Tokens}
    & \textbf{Prompt Compression} \\
    & \textbf{(all)} & & & & \textbf{(textual)} & \textbf{($\times$, $\uparrow$)} \\
    \midrule
    GLM-4-9B-Chat-1M & 10175.8 & 0.0 & 0.0 & 10175.8 & 80.4 & 1.00 \\
    Glyph-9B         & 174.8 & 3347.8 & 0.0 & 3522.5 & 427.4 & 3.14 \\
    \textbf{Seer (Ours)}$^{\dagger}$ & 4051.1 & 3572.7 & 3741.8 & 7623.8 & 577.1 & 1.28 \\
    \bottomrule
    \end{tabular}%
    }
    \caption{Token budget comparison on LongBench under a unified setup, reported as per-task means averaged over all 21 tasks. \textbf{Textual Prompt (all)} denotes all textual prompt tokens seen by the model; \textbf{Visual Prompt} and \textbf{Retrieved Text} denote stage-1 visual tokens and stage-2 retrieved textual tokens, respectively. \textbf{Total Prompt} is \textbf{Textual Prompt (all)} $+$ \textbf{Visual Prompt}; \textbf{Retrieved Text} is the retrieval-induced \emph{subset} of \textbf{Textual Prompt (all)} and is broken out separately rather than added again. Completion Tokens count generated textual outputs only. \textbf{Prompt Compression} ($\times$, $\uparrow$) is the mean over the 21 tasks of the per-task ratio \textit{GLM total prompt / model total prompt}, with GLM fixed to 1.00; because it averages per-task ratios rather than dividing the averaged totals, it does not equal the quotient of the Total Prompt column. Full per-task breakdown is reported in Appendix~\ref{app:token-breakdown}. $\dagger$ denotes our final model.}
    \label{tab:token-efficiency}
\end{table*}

\Cref{tab:token-efficiency} reports token usage under a unified setup.
Compared with GLM-4-9B-Chat-1M (10,175.8 average prompt tokens), Glyph-9B reduces prompt tokens to 3,522.5 on average (3.14$\times$ compression), while \textsc{Seer} uses 7,623.8 prompt tokens (1.28$\times$ compression) by selectively adding retrieved text only when needed.
This places \textsc{Seer} between full-text and pure visual compression: it sacrifices part of the maximal compression of Glyph-9B to recover task-relevant textual precision. We provide the full per-task token breakdown for all 21 LongBench tasks in Appendix~\ref{app:token-breakdown}.
Prompt-token counts are an architectural accounting quantity rather than a measured runtime; Appendix~\ref{app:wallclock} reports wall-clock measurements and scopes the efficiency claim to prompt and KV-cache footprint rather than to single-request latency.

\subsection{Ablation Study}
\label{sec:ablation-study}

\begin{table}[t]
    \centering
    \small
    \setlength{\tabcolsep}{7pt}
    \begin{tabular}{lcc}
    \toprule
    \textbf{Variant} & \textbf{LongBench Avg (\%)} & \textbf{$\Delta$ vs. Full} \\
    \midrule
    \textbf{\textsc{Seer} (Full)} & 51.11 & 0.00 \\
    \quad w/o Retrieve (Selection only) & 43.00 & $-8.11$ \\
    \quad w/o Selection + Retrieval (Glyph-9B) & 48.78 & $-2.33$ \\
    \quad w/o mechanism, same data (Glyph-9B-SFT-direct) & 45.20 & $-5.91$ \\
    \bottomrule
    \end{tabular}
    \caption{Ablation on LongBench using the overall category-level macro-average (across six major task categories). The Full variant is \textsc{Seer} with the complete Select-Retrieve-Reason pipeline. \textit{w/o Retrieve} keeps Stage-1 selection but disables Stage-2 text retrieval. \textit{w/o Selection + Retrieval} corresponds to Glyph-9B-style pure visual reasoning. \textit{w/o mechanism, same data} is an iso-training control trained on the same trajectories, initialization, optimizer, seed and step budget as \textsc{Seer}, with only the select/retrieve/observation turns removed (Appendix~\ref{app:iso-training}). $\Delta$ is computed as \textit{Avg(variant) $-$ Avg(full)}; negative values indicate performance degradation after removing components.}
    \label{tab:ablation-longbench}
\end{table}

We report ablations in a full-to-reduced order for direct comparison:
\textbf{full \textsc{Seer}} (complete Select-Retrieve-Reason), \textbf{w/o Retrieve} (selection-only, no retrieved text), and \textbf{w/o Selection + Retrieval} (Glyph-9B behavior), as summarized in \Cref{tab:ablation-longbench}.
The \textit{w/o Retrieve} variant obtains 43.00\% LongBench Avg ($\Delta=-8.11$), while \textit{w/o Selection + Retrieval} obtains 48.78\% ($\Delta=-2.33$). This non-monotonic trend suggests an interaction effect rather than an isolated gain from selection: Stage-1 selection primarily acts as a routing step for Stage-2 retrieval, and may provide limited standalone benefit when no retrieved text is provided. The result is consistent with our earlier observation that many failures of pure visual reasoning involve fine-grained textual details (e.g., numbers and Chinese strings), where explicit text evidence is often more reliable than OCR-like visual decoding alone.
The last row isolates the mechanism from the data: an iso-training control trained on the same trajectories but with the select/retrieve turns removed reaches 45.20, so fine-tuning on our data alone does not account for \textsc{Seer}'s accuracy (Appendix~\ref{app:iso-training}).
Two further controls support the same reading and are reported in the appendix: an intrinsic page-selection evaluation against newly annotated LongBench page labels, where \textsc{Seer} reaches 0.578 page F1 against 0.413 for random and 0.486 for first-page selection (Appendix~\ref{app:selection-quality}), and an end-to-end comparison in which only the page selector is swapped for BM25 (Appendix~\ref{app:bm25-downstream}).
Appendix~\ref{app:rlm} additionally compares \textsc{Seer} against a recursive text-side long-context reader on the LongBench QA tasks.

\subsection{Analysis of Selective Retrieval}
\label{sec:ablation-selection}

Unless otherwise noted, overall numbers in this subsection use the official LongBench category-level macro average (same aggregation as \cref{sec:ablation-study}).

\begin{figure*}[t]
    \centering
    \begin{subfigure}[t]{0.49\textwidth}
        \centering
        \includegraphics[width=\linewidth]{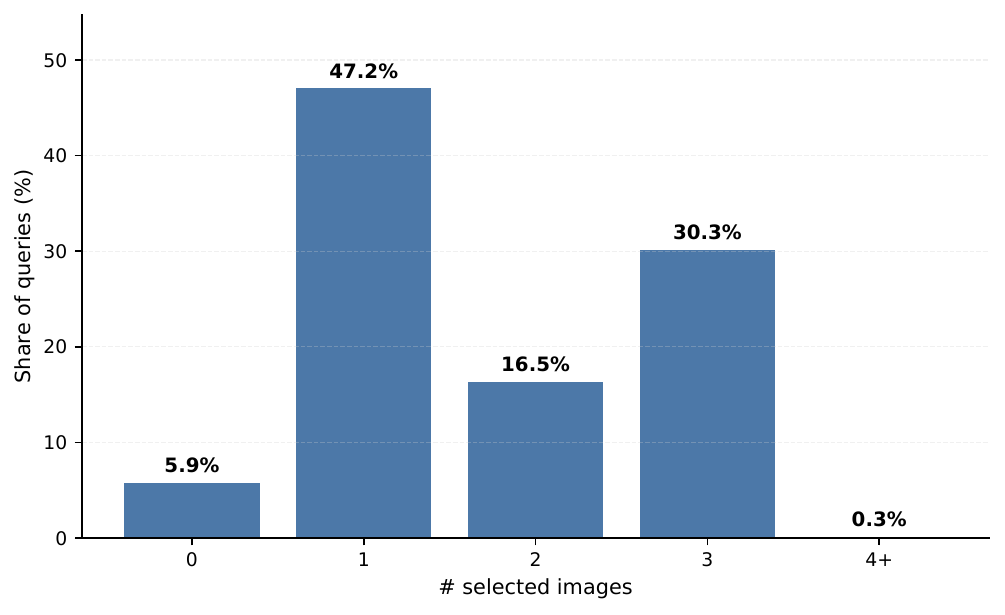}
        \caption{Overall selected-image histogram on LongBench.}
        \label{fig:selection_hist_longbench}
    \end{subfigure}
    \hfill
    \begin{subfigure}[t]{0.49\textwidth}
        \centering
        \includegraphics[width=\linewidth]{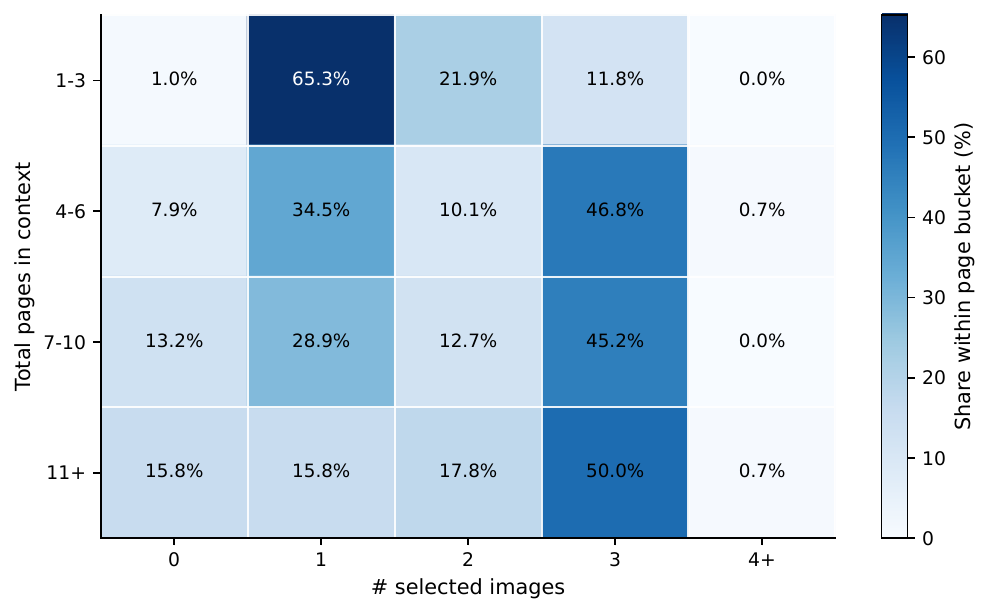}
        \caption{Selection distribution by context-length bucket (row-normalized).}
        \label{fig:selection_heatmap_page_bucket}
    \end{subfigure}
    \caption{Selective retrieval behavior on LongBench. \textbf{Left:} Distribution of selected images per query; counts here are the model's raw predicted selections prior to the canonicalization of \cref{sec:framework}. \textbf{Right:} For each total-page bucket (1--3, 4--6, 7--10, 11+), we show the fraction of samples selecting each image-count bucket (0, 1, 2, 3, 4+). The mode shifts from one to three images as contexts grow, but selections almost never exceed three: the budget saturates rather than scaling with context length.}
    \label{fig:selection_analysis_main}
\end{figure*}

\paragraph{How many images does \textsc{Seer} select?}
Across LongBench, \textsc{Seer} selects 1.59 \emph{valid} images per query on average (median 1), where valid means indices canonicalized to $[1,\texttt{total\_pages}]$ as described in \cref{sec:framework}; fewer than 1\% of queries select more than three images.
The selection is sparse relative to the available context, which averages 6.6 rendered pages per sample (31{,}182 pages over 4{,}750 examples; see Appendix~\ref{app:selection-quality}), consistent with the claim that only a small subset typically needs precise text retrieval.
As summarized in \Cref{fig:selection_analysis_main}, \Cref{fig:selection_hist_longbench} shows that the overall distribution is concentrated at low selected counts.
\Cref{fig:selection_heatmap_page_bucket} refines this picture: the modal selection shifts from one image on short contexts (1--3 pages) to three images on longer ones, but essentially no mass moves beyond three regardless of context length.
Selection therefore saturates at a small budget rather than scaling with the number of available pages.
This ceiling is inherited from the training distribution, and it is the mechanism behind the distributed-evidence limitation stated in the Limitations section: when the evidence spans more than three pages, a single selection round is recall-limited by construction.

\begin{figure*}[t]
    \centering
    \includegraphics[width=1\textwidth]{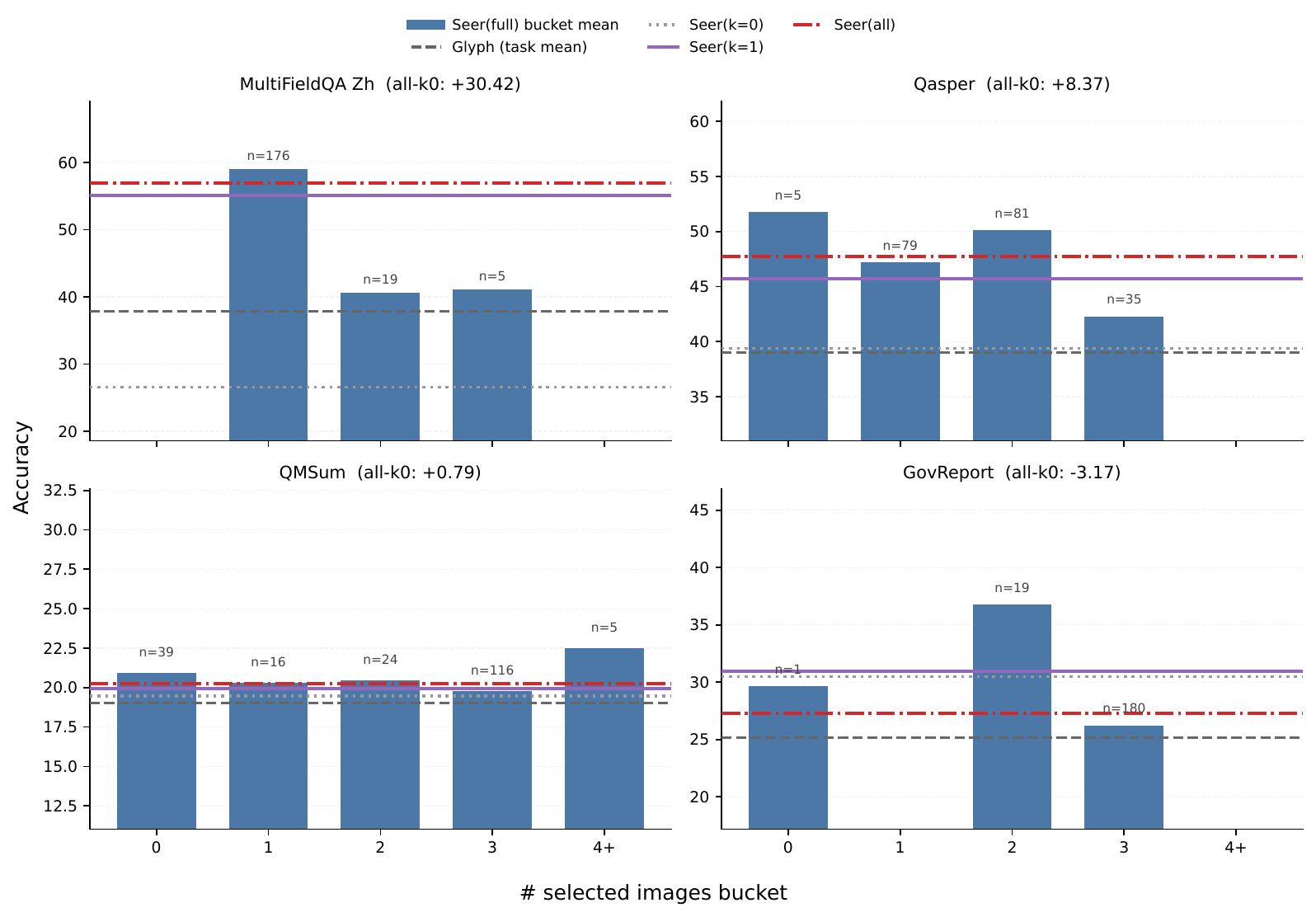}
    \caption{Representative-task analysis of retrieval effectiveness by selected-image bucket. Blue bars show Seer(full) bucket-level mean accuracy for selected-count buckets (0,1,2,3,4+), with per-bucket sample counts annotated as $n$. Horizontal reference lines indicate task-level mean scores of \textit{Glyph}, \textit{Seer(k=0)}, \textit{Seer(k=1)}, and \textit{Seer(all)}. The figure highlights task heterogeneity: retrieval is beneficial on some tasks but not uniformly so across all tasks.}
    \label{fig:selection_bucket_representative_tasks}
\end{figure*}

\paragraph{How does retrieval scope affect accuracy?}
We evaluate retrieval effectiveness using endpoint ablation, where $k{=}0$ corresponds to ``w/o Retrieve'' (Appendix~\ref{app:wo-retrieve}) and $k{=}\text{all}$ corresponds to full retrieval.
For consistency with the main results, endpoint comparison uses the primary ablation setting: the category-level macro average improves from 43.00 ($k{=}0$) to 51.11 ($k{=}\text{all}$), i.e., +8.11.
We additionally observe consistent recovery at an intermediate retrieval budget ($k{=}1$).
This comparison is computed under the task-level (unweighted-over-tasks) summary used for the bucket analysis rather than the category-level macro average, so its $k{=}0\!\to\!k{=}\text{all}$ endpoint gap is $+4.88$ rather than $+8.11$; of that gap, $k{=}1$ alone already recovers $+3.00$, i.e., 61.5\%.
Category-level breakdown under the same summary shows the strongest early gain on single-document QA (+13.73 from $k{=}0$ to $k{=}1$, versus +1.25 from $k{=}1$ to $k{=}\text{all}$), consistent with our extraction-focused hypothesis.
At task level, we report bucketed mean accuracies by selected-image count to visualize heterogeneity: gains are large on some tasks (e.g., MultiFieldQA-Zh, Qasper) but limited or negative on others (e.g., QMSum, GovReport).
To show this heterogeneity, \Cref{fig:selection_bucket_representative_tasks} presents representative tasks spanning strong-positive, mild-positive, near-neutral, and negative cases.
More results are reported in Appendix~\ref{app:selection-bucket-alltasks} (\Cref{fig:selection_bucket_alltasks_app}).

\subsection{Error Analysis}

We provide a compact error analysis focused on the Chinese-language tasks, motivated by our earlier observation that densely rendered Chinese amplifies the cost of purely visual decoding.
\begin{table}[t]
    \centering
    \small
    \setlength{\tabcolsep}{8pt}
    \begin{tabular}{lccc}
    \toprule
    \textbf{Task (Chinese)} & \textbf{Glyph-9B} & \textbf{\textsc{Seer}} & \textbf{$\Delta$} \\
    \midrule
    MultiFieldQA-Zh & 37.82 & 56.99 & $+19.17$ \\
    PassageRetrieval-Zh & 91.00 & 97.00 & $+6.00$ \\
    VCSUM & 12.28 & 14.20 & $+1.92$ \\
    \midrule
    Mean (3 tasks) & 47.03 & 56.06 & $+9.03$ \\
    \bottomrule
    \end{tabular}
    \caption{Error-analysis summary on the three Chinese-language LongBench tasks. Values are task scores (\%) under official LongBench metrics. $\Delta$ is computed as \textit{\textsc{Seer} $-$ Glyph-9B}.}
    \label{tab:error-cjk}
\end{table}

As shown in \Cref{tab:error-cjk}, \textsc{Seer} shows gains over Glyph-9B on all three Chinese tasks, with especially large gains on MultiFieldQA-Zh (+19.17) and strong gains on PassageRetrieval-Zh (+6.00).
The smaller improvement on VCSUM (+1.92) suggests that retrieval may be more beneficial for extraction-heavy settings, while abstractive summarization remains more challenging.
Appendix~\ref{app:qualitative} gives representative success and failure cases drawn from the prediction files.

\section{Conclusion}

Visual-text compression buys efficiency by treating every page alike, which is what costs it precision on questions that hinge on a few exact details.
\textsc{Seer} makes that compression query-selective: it reads the whole rendered document, identifies the pages a query actually depends on, and retrieves exact source text only there.
On LongBench this reaches 51.11\% overall accuracy, 2.33 points above the Glyph-9B model it starts from, with the gains concentrated on extraction-heavy tasks; our ablations and controls attribute the improvement to the select-retrieve mechanism rather than to fine-tuning on our trajectories.
The selection the model learns is sparse, 1.59 pages per query on average.
This is why a small retrieval budget already recovers most of the gain, and also why questions whose evidence is scattered across many pages remain the hardest case.
\textsc{Seer} therefore trades part of the token compression for precision where it is needed; making the retrieval budget adaptive, and iterating the selection rather than committing to it in one shot, is the natural next step.

\section*{Limitations}
Four caveats bound what the results above establish.
First, selection is not robust as a deployment interface: the SFT model occasionally emits
out-of-range page indices, which we discard rather than correct, and which would propagate to
retrieval failures in a system that trusted them.
Second, the efficiency claim is narrower than the compression ratio suggests. The token savings are
an average, not a guarantee: on seven of the 21 tasks \textsc{Seer}'s prompt is larger than the
full-text prompt, because retrieval overhead outweighs what rendering saves. Where savings do
materialize, they are in prompt and KV-cache footprint rather than in speed. At batch size 1 on
LongBench contexts, \textsc{Seer}'s median end-to-end latency is 5.47\,s against 1.07\,s for
full-text inference, because it emits an explicit reasoning trace where the baseline answers in a
handful of tokens (Appendix~\ref{app:wallclock}).
Third, the accuracy gain is specific to questions whose evidence sits on a few pages, and it thins
out as evidence spreads. Our answer-conditioned teacher filtering also underrepresents that regime
at training time, so the limitation is partly built into the data.
Fourth, \textsc{Seer} is specialized for question answering over rendered long documents: we did not
evaluate it on natural images, charts, or OCR benchmarks and make no claim there.
The remedies we consider most promising are constrained decoding with index validation, an adaptive
rather than fixed retrieval budget, iterative selection for distributed evidence, and inference-time
optimization such as tree-based decoding~\citep{xu2025dts}.
None of these caveats touch the central comparison, which is same-family and held fixed on every
axis but the mechanism: selective retrieval is worth a measurable accuracy gain over uniform visual
compression on extraction-heavy long-context tasks.

\section*{Acknowledgments}
This work was supported in part by the NSF AI Institute for Foundations of Machine Learning (IFML) at UT Austin, the NIH Common Fund Bridge2AI program (CM4AI; grant 1OT2OD032742-01), and the National Science Foundation (grants 2303038 and IIS-2551752).

\bibliography{colm2026visual}

\begin{thebibliography}{33}
\providecommand{\natexlab}[1]{#1}
\providecommand{\url}[1]{\texttt{#1}}
\expandafter\ifx\csname urlstyle\endcsname\relax
  \providecommand{\doi}[1]{doi: #1}\else
  \providecommand{\doi}{doi: \begingroup \urlstyle{rm}\Url}\fi

\bibitem[Bai et~al.(2023)Bai, Bai, Yang, Wang, Tan, Wang, Lin, Zhou, and Zhou]{bai2023qwenvl}
Jinze Bai, Shuai Bai, Shusheng Yang, Shijie Wang, Sinan Tan, Peng Wang, Junyang Lin, Chang Zhou, and Jingren Zhou.
\newblock {Qwen-VL}: A versatile vision-language model for understanding, localization, text reading, and beyond.
\newblock \emph{arXiv preprint arXiv:2308.12966}, 2023.

\bibitem[Bai et~al.(2024)Bai, Lv, Zhang, Lyu, Tang, Huang, Du, Liu, Zeng, Hou, et~al.]{bai2024longbench}
Yushi Bai, Xin Lv, Jiajie Zhang, Hongchang Lyu, Jiankai Tang, Zhidian Huang, Zhengxiao Du, Xiao Liu, Aohan Zeng, Lei Hou, et~al.
\newblock {LongBench}: A bilingual, multitask benchmark for long context understanding.
\newblock In \emph{Proceedings of the 62nd Annual Meeting of the Association for Computational Linguistics (Volume 1: Long Papers)}, pp.\  3119--3137, 2024.

\bibitem[Chen et~al.(2024)Chen, Wu, Wang, Su, Chen, Xing, Zhong, Zhang, Zhu, Lu, Li, Luo, Lu, Qiao, and Dai]{chen2024internvl}
Zhe Chen, Jiannan Wu, Wenhai Wang, Weijie Su, Guo Chen, Sen Xing, Muyan Zhong, Qinglong Zhang, Xizhou Zhu, Lewei Lu, Bin Li, Ping Luo, Tong Lu, Yu~Qiao, and Jifeng Dai.
\newblock {InternVL}: Scaling up vision foundation models and aligning for generic visual-linguistic tasks.
\newblock In \emph{Proceedings of the IEEE/CVF Conference on Computer Vision and Pattern Recognition (CVPR)}, pp.\  24185--24198, 2024.

\bibitem[Cheng et~al.(2026)Cheng, Liu, Zhang, Fei, Hong, Lyu, Wang, Su, Gu, Liu, Bai, Tang, Wang, and Huang]{cheng2026glyph}
Jiale Cheng, Yusen Liu, Xinyu Zhang, Yulin Fei, Wenyi Hong, Ruiliang Lyu, Weihan Wang, Zhe Su, Xiaotao Gu, Xiao Liu, Yushi Bai, Jie Tang, Hongning Wang, and Minlie Huang.
\newblock Glyph: Scaling context windows via visual-text compression.
\newblock In \emph{Proceedings of the 64th Annual Meeting of the {A}ssociation for {C}omputational {L}inguistics (Volume 1: Long Papers)}, pp.\  37145--37158, San Diego, California, United States, July 2026. Association for Computational Linguistics.

\bibitem[Deng et~al.(2025)Deng, Da, Pan, He, Ide, Garg, Lauffer, Park, Pasari, Rane, et~al.]{deng2025swe}
Xiang Deng, Jeff Da, Edwin Pan, Yannis~Yiming He, Charles Ide, Kanak Garg, Niklas Lauffer, Andrew Park, Nitin Pasari, Chetan Rane, et~al.
\newblock {SWE}-bench {P}ro: Can {AI} agents solve long-horizon software engineering tasks?
\newblock \emph{arXiv preprint arXiv:2509.16941}, 2025.

\bibitem[Glm et~al.(2024)Glm, Zeng, Xu, Wang, Zhang, Yin, Zhang, Rojas, Feng, Zhao, et~al.]{glm2024chatglm}
Team Glm, Aohan Zeng, Bin Xu, Bowen Wang, Chenhui Zhang, Da~Yin, Dan Zhang, Diego Rojas, Guanyu Feng, Hanlin Zhao, et~al.
\newblock {ChatGLM}: A family of large language models from {GLM}-130{B} to {GLM}-4 all tools.
\newblock \emph{arXiv preprint arXiv:2406.12793}, 2024.

\bibitem[Grattafiori et~al.(2024)Grattafiori, Dubey, Jauhri, Pandey, Kadian, Al-Dahle, Letman, Mathur, Schelten, Vaughan, et~al.]{grattafiori2024llama}
Aaron Grattafiori, Abhimanyu Dubey, Abhinav Jauhri, Abhinav Pandey, Abhishek Kadian, Ahmad Al-Dahle, Aiesha Letman, Akhil Mathur, Alan Schelten, Alex Vaughan, et~al.
\newblock The {L}lama 3 herd of models.
\newblock \emph{arXiv preprint arXiv:2407.21783}, 2024.

\bibitem[Guo et~al.(2025)Guo, Yang, Zhang, Song, Wang, Zhu, Xu, Zhang, Ma, Bi, et~al.]{guo2025deepseek}
Daya Guo, Dejian Yang, Haowei Zhang, Junxiao Song, Peiyi Wang, Qihao Zhu, Runxin Xu, Ruoyu Zhang, Shirong Ma, Xiao Bi, et~al.
\newblock {DeepSeek-R1}: Incentivizing reasoning capability in {LLM}s via reinforcement learning.
\newblock \emph{arXiv preprint arXiv:2501.12948}, 2025.

\bibitem[Jaech et~al.(2024)Jaech, Kalai, Lerer, Richardson, El-Kishky, Low, Helyar, Madry, Beutel, Carney, et~al.]{jaech2024openai}
Aaron Jaech, Adam Kalai, Adam Lerer, Adam Richardson, Ahmed El-Kishky, Aiden Low, Alec Helyar, Aleksander Madry, Alex Beutel, Alex Carney, et~al.
\newblock {OpenAI} o1 system card.
\newblock \emph{arXiv preprint arXiv:2412.16720}, 2024.

\bibitem[Jiang et~al.(2024)Jiang, Wu, Luo, Li, Lin, Yang, and Qiu]{jiang2024longllmlingua}
Huiqiang Jiang, Qianhui Wu, Xufang Luo, Dongsheng Li, Chin-Yew Lin, Yuqing Yang, and Lili Qiu.
\newblock {LongLLMLingua}: Accelerating and enhancing {LLM}s in long context scenarios via prompt compression.
\newblock In \emph{Proceedings of the 62nd Annual Meeting of the Association for Computational Linguistics (Volume 1: Long Papers)}, pp.\  1658--1677, 2024.

\bibitem[Liu et~al.(2024)Liu, Lin, Hewitt, Paranjape, Bevilacqua, Petroni, and Liang]{liu2024lost}
Nelson~F Liu, Kevin Lin, John Hewitt, Ashwin Paranjape, Michele Bevilacqua, Fabio Petroni, and Percy Liang.
\newblock Lost in the middle: How language models use long contexts.
\newblock \emph{Transactions of the Association for Computational Linguistics}, 12:\penalty0 157--173, 2024.

\bibitem[Luo et~al.(2026{\natexlab{a}})Luo, Chuang, Wang, Le, Zhong, Liu, Yuan, Sui, Braverman, Chaudhary, et~al.]{luo2026autol2s}
Feng Luo, Yu-Neng Chuang, Guanchu Wang, Hoang Anh~Duy Le, Shaochen Zhong, Hongyi Liu, Jiayi Yuan, Yang Sui, Vladimir Braverman, Vipin Chaudhary, et~al.
\newblock {AutoL2S}: Auto long-short reasoning for efficient large language models.
\newblock In \emph{Findings of the Association for Computational Linguistics: ACL 2026}, pp.\  16836--16858, 2026{\natexlab{a}}.

\bibitem[Luo et~al.(2026{\natexlab{b}})Luo, Chuang, Wang, Xu, Han, Zhang, and Braverman]{luo2026demystifying}
Feng Luo, Yu-Neng Chuang, Guanchu Wang, Zicheng Xu, Xiaotian Han, Tianyi Zhang, and Vladimir Braverman.
\newblock Demystifying {OPD}: Length inflation and stabilization strategies for large language models.
\newblock \emph{arXiv preprint arXiv:2604.08527}, 2026{\natexlab{b}}.

\bibitem[Maharana et~al.(2024)Maharana, Lee, Tulyakov, Bansal, Barbieri, and Fang]{maharana2024evaluating}
Adyasha Maharana, Dong-Ho Lee, Sergey Tulyakov, Mohit Bansal, Francesco Barbieri, and Yuwei Fang.
\newblock Evaluating very long-term conversational memory of {LLM} agents.
\newblock In \emph{Proceedings of the 62nd Annual Meeting of the Association for Computational Linguistics (Volume 1: Long Papers)}, pp.\  13851--13870, 2024.

\bibitem[Merrill et~al.(2026)Merrill, Shaw, Carlini, Li, Raj, Bercovich, Shi, Shin, Walshe, Buchanan, Shen, Ye, Lin, Poulos, Wang, Jitsev, Nezhurina, Lu, Mastromichalakis, Xu, Chen, Liu, Zhang, Chen, Kashyap, Uslu, Li, Wu, Yan, Bian, Sharma, Sun, Dillmann, Anand, Lanpouthakoun, Koopah, Hu, Guha, Dreiman, Zhu, Krauth, Zhong, Muennighoff, Amanfu, Tan, Pimpalgaonkar, Aggarwal, Lin, Lan, Zhao, Liang, Wang, Wang, Zhou, Heineman, Liu, Trivedi, Yang, Lin, Shetty, Yang, Omi, Raoof, Li, Zhuo, Lin, Dai, Wang, Chai, Zhou, Wahdany, She, Hu, Dong, Zhu, Cui, Saiyed, Kolbeinsson, Rytting, Marten, Wang, Dimakis, Konwinski, and Schmidt]{merrill2026terminalbench}
Mike~A Merrill, Alexander~Glenn Shaw, Nicholas Carlini, Boxuan Li, Harsh Raj, Ivan Bercovich, Lin Shi, Jeong~Yeon Shin, Thomas Walshe, E.~Kelly Buchanan, Junhong Shen, Guanghao Ye, Haowei Lin, Jason Poulos, Maoyu Wang, Jenia Jitsev, Marianna Nezhurina, Di~Lu, Orfeas~Menis Mastromichalakis, Zhiwei Xu, Zizhao Chen, Yue Liu, Robert Zhang, Leon~Liangyu Chen, Anurag Kashyap, Jan-Lucas Uslu, Jeffrey Li, Jianbo Wu, Minghao Yan, Song Bian, Vedang Sharma, Ke~Sun, Steven Dillmann, Akshay Anand, Andrew Lanpouthakoun, Bardia Koopah, Changran Hu, Etash~Kumar Guha, Gabriel H.~S. Dreiman, Jiacheng Zhu, Karl Krauth, Li~Zhong, Niklas Muennighoff, Robert~Kwesi Amanfu, Shangyin Tan, Shreyas Pimpalgaonkar, Tushar Aggarwal, Xiangning Lin, Xin Lan, Xuandong Zhao, Yiqing Liang, Yuanli Wang, Zilong Wang, Changzhi Zhou, David Heineman, Hange Liu, Harsh Trivedi, John Yang, Junhong Lin, Manish Shetty, Michael Yang, Nabil Omi, Negin Raoof, Shanda Li, Terry~Yue Zhuo, Wuwei Lin, Yiwei Dai, Yuxin Wang, Wenhao Chai, Shang Zhou, Dariush
  Wahdany, Ziyu She, Jiaming Hu, Zhikang Dong, Yuxuan Zhu, Sasha Cui, Ahson Saiyed, Arinbj{\"o}rn Kolbeinsson, Christopher~Michael Rytting, Ryan Marten, Yixin Wang, Alex Dimakis, Andy Konwinski, and Ludwig Schmidt.
\newblock {Terminal-Bench}: Benchmarking agents on hard, realistic tasks in command line interfaces.
\newblock In \emph{The Fourteenth International Conference on Learning Representations}, 2026.
\newblock URL \url{https://openreview.net/forum?id=a7Qa4CcHak}.

\bibitem[Peng et~al.(2024)Peng, Quesnelle, Fan, and Shippole]{peng2024yarn}
Bowen Peng, Jeffrey Quesnelle, Honglu Fan, and Enrico Shippole.
\newblock {YaRN}: Efficient context window extension of large language models.
\newblock In \emph{The Twelfth International Conference on Learning Representations}, 2024.
\newblock URL \url{https://openreview.net/forum?id=wHBfxhZu1u}.

\bibitem[Qian et~al.(2025)Qian, Acikgoz, He, Wang, Chen, Hakkani-T{\"u}r, Tur, and Ji]{qian2025toolrl}
Cheng Qian, Emre~Can Acikgoz, Qi~He, Hongru Wang, Xiusi Chen, Dilek Hakkani-T{\"u}r, Gokhan Tur, and Heng Ji.
\newblock {ToolRL}: Reward is all tool learning needs.
\newblock \emph{arXiv preprint arXiv:2504.13958}, 2025.

\bibitem[Robertson \& Zaragoza(2009)Robertson and Zaragoza]{robertson2009probabilistic}
Stephen Robertson and Hugo Zaragoza.
\newblock The probabilistic relevance framework: {BM25} and beyond.
\newblock \emph{Foundations and Trends in Information Retrieval}, 3\penalty0 (4):\penalty0 333--389, 2009.

\bibitem[Sui et~al.(2025)Sui, Chuang, Wang, Zhang, Zhang, Yuan, Liu, Wen, Zhong, Zou, et~al.]{sui2025stop}
Yang Sui, Yu-Neng Chuang, Guanchu Wang, Jiamu Zhang, Tianyi Zhang, Jiayi Yuan, Hongyi Liu, Andrew Wen, Shaochen Zhong, Na~Zou, et~al.
\newblock Stop overthinking: A survey on efficient reasoning for large language models.
\newblock \emph{arXiv preprint arXiv:2503.16419}, 2025.

\bibitem[Sumers et~al.(2024)Sumers, Yao, Narasimhan, and Griffiths]{sumers2024cognitive}
Theodore Sumers, Shunyu Yao, Karthik~R Narasimhan, and Thomas~L. Griffiths.
\newblock Cognitive architectures for language agents.
\newblock \emph{Transactions on Machine Learning Research}, 2024.
\newblock ISSN 2835-8856.
\newblock URL \url{https://openreview.net/forum?id=1i6ZCvflQJ}.
\newblock Survey Certification, Featured Certification.

\bibitem[Trivedi et~al.(2023)Trivedi, Balasubramanian, Khot, and Sabharwal]{trivedi2023interleaving}
Harsh Trivedi, Niranjan Balasubramanian, Tushar Khot, and Ashish Sabharwal.
\newblock Interleaving retrieval with chain-of-thought reasoning for knowledge-intensive multi-step questions.
\newblock In \emph{Proceedings of the 61st Annual Meeting of the Association for Computational Linguistics (Volume 1: Long Papers)}, pp.\  10014--10037, 2023.

\bibitem[Vaswani et~al.(2017)Vaswani, Shazeer, Parmar, Uszkoreit, Jones, Gomez, Kaiser, and Polosukhin]{vaswani2017attention}
Ashish Vaswani, Noam Shazeer, Niki Parmar, Jakob Uszkoreit, Llion Jones, Aidan~N Gomez, {\L}ukasz Kaiser, and Illia Polosukhin.
\newblock Attention is all you need.
\newblock \emph{Advances in Neural Information Processing Systems}, 30, 2017.

\bibitem[Wei et~al.(2025)Wei, Sun, and Li]{wei2025deepseek}
Haoran Wei, Yaofeng Sun, and Yukun Li.
\newblock {DeepSeek-OCR}: Contexts optical compression.
\newblock \emph{arXiv preprint arXiv:2510.18234}, 2025.

\bibitem[Wei et~al.(2022)Wei, Wang, Schuurmans, Bosma, Xia, Chi, Le, Zhou, et~al.]{wei2022chain}
Jason Wei, Xuezhi Wang, Dale Schuurmans, Maarten Bosma, Fei Xia, Ed~Chi, Quoc~V Le, Denny Zhou, et~al.
\newblock Chain-of-thought prompting elicits reasoning in large language models.
\newblock \emph{Advances in Neural Information Processing Systems}, 35:\penalty0 24824--24837, 2022.

\bibitem[Wu et~al.(2024)Wu, Gu, Feng, Zhong, Xu, Yang, Liu, and Qin]{wu2024extending}
Yingsheng Wu, Yuxuan Gu, Xiaocheng Feng, Weihong Zhong, Dongliang Xu, Qing Yang, Hongtao Liu, and Bing Qin.
\newblock Extending context window of large language models from a distributional perspective.
\newblock In \emph{Proceedings of the 2024 Conference on Empirical Methods in Natural Language Processing}, pp.\  7288--7301, 2024.

\bibitem[Xing et~al.(2025)Xing, Wang, Yan, Shu, and Tang]{xing2025visioncentric}
Ling Xing, Alex~Jinpeng Wang, Rui Yan, Xiangbo Shu, and Jinhui Tang.
\newblock Vision-centric token compression in large language model.
\newblock In \emph{The Thirty-ninth Annual Conference on Neural Information Processing Systems}, 2025.
\newblock URL \url{https://openreview.net/forum?id=YdggdEL41C}.

\bibitem[Xu et~al.(2025)Xu, Lou, Wang, Chuang, Luo, Zheng, Szalay, Liu, and Braverman]{xu2025dts}
Zicheng Xu, Xiuyi Lou, Guanchu Wang, Yu-Neng Chuang, Feng Luo, Guangyao Zheng, Alexander~S Szalay, Zirui Liu, and Vladimir Braverman.
\newblock {DTS}: Enhancing large reasoning models via decoding tree sketching.
\newblock \emph{arXiv preprint arXiv:2511.00640}, 2025.

\bibitem[Yang et~al.(2025{\natexlab{a}})Yang, Li, Yang, Zhang, Hui, Zheng, Yu, Gao, Huang, Lv, et~al.]{yang2025qwen3}
An~Yang, Anfeng Li, Baosong Yang, Beichen Zhang, Binyuan Hui, Bo~Zheng, Bowen Yu, Chang Gao, Chengen Huang, Chenxu Lv, et~al.
\newblock {Q}wen3 technical report.
\newblock \emph{arXiv preprint arXiv:2505.09388}, 2025{\natexlab{a}}.

\bibitem[Yang et~al.(2025{\natexlab{b}})Yang, Yu, Li, Liu, Huang, Huang, Jiang, Tu, Zhang, Zhou, et~al.]{yang2025qwen25_1m}
An~Yang, Bowen Yu, Chengyuan Li, Dayiheng Liu, Fei Huang, Haoyan Huang, Jiandong Jiang, Jianhong Tu, Jianwei Zhang, Jingren Zhou, et~al.
\newblock {Q}wen2.5-1{M} technical report.
\newblock \emph{arXiv preprint arXiv:2501.15383}, 2025{\natexlab{b}}.

\bibitem[Yang et~al.(2024)Yang, Wang, Shen, Panda, and Kim]{yang2024gated}
Songlin Yang, Bailin Wang, Yikang Shen, Rameswar Panda, and Yoon Kim.
\newblock Gated linear attention transformers with hardware-efficient training.
\newblock In \emph{Proceedings of the 41st International Conference on Machine Learning}, ICML'24. JMLR.org, 2024.

\bibitem[Yao et~al.(2022)Yao, Zhao, Yu, Du, Shafran, Narasimhan, and Cao]{yao2022react}
Shunyu Yao, Jeffrey Zhao, Dian Yu, Nan Du, Izhak Shafran, Karthik~R Narasimhan, and Yuan Cao.
\newblock {ReAct}: Synergizing reasoning and acting in language models.
\newblock In \emph{The Eleventh International Conference on Learning Representations}, 2022.

\bibitem[Yuan et~al.(2024)Yuan, Liu, Zhong, Chuang, Li, Wang, Le, Jin, Chaudhary, Xu, et~al.]{yuan2024kv}
Jiayi Yuan, Hongyi Liu, Shaochen Zhong, Yu-Neng Chuang, Songchen Li, Guanchu Wang, Duy Le, Hongye Jin, Vipin Chaudhary, Zhaozhuo Xu, et~al.
\newblock {KV} cache compression, but what must we give in return? a comprehensive benchmark of long context capable approaches.
\newblock In \emph{Findings of the Association for Computational Linguistics: EMNLP 2024}, pp.\  4623--4648, 2024.

\bibitem[Zhang et~al.(2025)Zhang, Kraska, and Khattab]{zhang2025recursive}
Alex~L Zhang, Tim Kraska, and Omar Khattab.
\newblock Recursive language models.
\newblock \emph{arXiv preprint arXiv:2512.24601}, 2025.

\end{thebibliography}
\bibliographystyle{colm2026_conference}

\clearpage
\appendix
\section{Appendix}
\makeatletter
\setlength{\@fptop}{0pt}
\makeatother


\subsection{Implementation Details}
\label{app:implementation}

\subsubsection{Model Architecture}

We build upon Glyph-9B, which is based on the GLM-4V architecture (\texttt{Glm4vForConditionalGeneration}).
The model consists of a vision encoder and a language model backbone; detailed specifications are listed in \Cref{tab:model-arch}.

\begin{table}[h]
\centering
\small
\begin{tabular}{ll}
\toprule
\textbf{Component} & \textbf{Specification} \\
\midrule
\multicolumn{2}{l}{\textit{Language Model}} \\
Hidden size & 4096 \\
Intermediate size & 13696 \\
Number of layers & 40 \\
Attention heads & 32 (2 KV heads) \\
Max position embeddings & 131072 \\
Vocabulary size & 151552 \\
\midrule
\multicolumn{2}{l}{\textit{Vision Encoder}} \\
Hidden size & 1536 \\
Depth & 24 \\
Attention heads & 12 \\
Image size & 336 \\
Patch size & 14 \\
Spatial merge size & 2 \\
\bottomrule
\end{tabular}
\caption{Model architecture specifications.}
\label{tab:model-arch}
\end{table}

\subsubsection{Special Tokens}
\label{app:special-tokens}

We leverage the special tokens already present in the Glyph tokenizer.
\Cref{tab:special-tokens} lists the tokens used in our framework and their corresponding IDs.

\begin{table}[h]
\centering
\small
\begin{tabular}{llc}
\toprule
\textbf{Token} & \textbf{Purpose} & \textbf{ID} \\
\midrule
\multicolumn{3}{l}{\textit{Conversation Structure}} \\
\texttt{<|endoftext|{}>} & End of text / Padding & 151329 \\
\texttt{<|user|{}>} & User turn marker & 151336 \\
\texttt{<|assistant|{}>} & Assistant turn marker & 151337 \\
\texttt{<|observation|{}>} & Tool response marker & 151338 \\
\midrule
\multicolumn{3}{l}{\textit{Reasoning and Tool Use}} \\
\texttt{<think>} & Start of reasoning & 151350 \\
\texttt{</think>} & End of reasoning & 151351 \\
\texttt{<tool\_call>} & Start of tool call & 151352 \\
\texttt{</tool\_call>} & End of tool call & 151353 \\
\midrule
\multicolumn{3}{l}{\textit{Vision}} \\
\texttt{<|begin\_of\_image|{}>} & Image start & 151339 \\
\texttt{<|end\_of\_image|{}>} & Image end & 151340 \\
\texttt{<|image|{}>} & Image placeholder & 151363 \\
\bottomrule
\end{tabular}
\caption{Special tokens used in our framework and their token IDs.}
\label{tab:special-tokens}
\end{table}

\paragraph{EOS Token Handling.}
The model's EOS token list includes three tokens: 151329 (\texttt{<|endoftext|{}>}), 151336 (\texttt{<|user|{}>}), and 151338 (\texttt{<|observation|{}>}).
During training and inference, we configure the generation to:
\begin{itemize}
    \item Stop at \texttt{</tool\_call>} (ID 151353) to execute the tool
    \item \textbf{Not} stop at \texttt{<|observation|{}>} (ID 151338), allowing generation to continue after tool response
    \item Stop at \texttt{<|user|{}>} (ID 151336) for final termination
\end{itemize}

\subsubsection{Data Format Examples}
\label{app:data-format}

\paragraph{Image Index Markers.}
Since the model receives images as sequences of vision tokens (\texttt{<|begin\_of\_image|>...<|end\_of\_image|>}), it cannot inherently distinguish between different images by index.
To enable the model to reference specific images in its selection output, we prepend a textual index marker before each image:

{\small
\begin{verbatim}
[Image 1]
<|begin_of_image|><|image|><|end_of_image|>
[Image 2]
<|begin_of_image|><|image|><|end_of_image|>
...
\end{verbatim}
}

This is standard practice for multi-image VLMs, e.g., Qwen-VL~\citep{bai2023qwenvl}, and InternVL~\citep{chen2024internvl}.
When using vLLM's chat API, this is achieved by interleaving text and image content blocks:
\begin{quote}
\small
\ttfamily
content = {[}\\
~~~~\{"type": "text", "text": "{[}Image 1{]}"\},\\
~~~~\{"type": "image\_url", "image\_url": \{"url": "img1.png"\}\},\\
~~~~\{"type": "text", "text": "{[}Image 2{]}"\},\\
~~~~\{"type": "image\_url", "image\_url": \{"url": "img2.png"\}\},\\
~~~~...\\
{]}
\end{quote}

The model can then output selection results by referencing these indices, e.g., \texttt{\{"indices": [1, 3]\}}.

\paragraph{Complete Trajectory Example.}
Below is a concrete example of a complete training trajectory:

\begin{quote}
\small
\ttfamily
<|user|{}>\\
{[}Image 1{]} <|begin\_of\_image|{}><|image|{}><|end\_of\_image|{}>\\
{[}Image 2{]} <|begin\_of\_image|{}><|image|{}><|end\_of\_image|{}>\\
{[}Image 3{]} <|begin\_of\_image|{}><|image|{}><|end\_of\_image|{}>\\
Query: What is the total price of the laptop and the monitor?\\
<|assistant|{}>\\
<tool\_call>\{"name": "select", "indices": [1, 3]\}</tool\_call>\\
<|observation|{}>\\
Text from image 1: Product: Dell XPS 15 Laptop. Price: \$1,299.99. Features: 15.6" display, Intel i7...\\
Text from image 3: Product: LG UltraWide Monitor. Price: \$449.99. Features: 34" curved display...\\
<think>\\
From the retrieved text, I can see that:\\
- The laptop (Dell XPS 15) costs \$1,299.99\\
- The monitor (LG UltraWide) costs \$449.99\\
Total: \$1,299.99 + \$449.99 = \$1,749.98\\
</think>\\
The total price is \$1,749.98.
\end{quote}

\paragraph{Trajectory-Level Loss Mask Example.}
For the trajectory above, the token-level mask assignment is:
\begin{itemize}
    \item \texttt{<|user|{}>} through the query: \textbf{mask = 0} (input, not trained)
    \item \texttt{<|assistant|>}\textbackslash n: \textbf{mask = 0} (generation prompt, not trained)
    \item \texttt{<tool\_call>}...\texttt{</tool\_call>}: \textbf{mask = 1} (model-generated, trained)
    \item \texttt{<|observation|{}>} through retrieved text: \textbf{mask = 0} (system-provided, not trained)
    \item \texttt{<think>}...\texttt{</think>}: \textbf{mask = 1} (model-generated, trained)
    \item Final answer: \textbf{mask = 1} (model-generated, trained)
\end{itemize}

\subsection{Training Data Construction}
\label{app:data-construction}

\paragraph{Data Sources.}
We construct training data from public long-context QA and document understanding datasets, ensuring no overlap with LongBench's test data.
The processing pipeline contains four stages: Download (raw), Render, Phase1 kept (\texttt{relevant\_indices} non-empty), and Phase2 final filtered (used for training in this paper).
\Cref{tab:data-pipeline-overview} summarizes the global retention at each stage across all 10 downloaded datasets.

\begin{table}[h]
    \centering
    \small
    \begin{tabular}{lrr}
    \toprule
    \textbf{Stage} & \textbf{\# Datasets} & \textbf{\# Samples} \\
    \midrule
    Download (raw) & 10 & 90,777 \\
    Render & 10 & 38,577 \\
    Phase1 kept (\texttt{relevant\_indices} $\neq \emptyset$) & 10 & 30,697 \\
    Phase2 final filtered (used) & 7 & 10,041 \\
    \bottomrule
    \end{tabular}
    \caption{Dataset and sample retention across the data construction pipeline.}
    \label{tab:data-pipeline-overview}
\end{table}

\paragraph{Per-Dataset Retention.}
\Cref{tab:training-data} reports sample counts for each included training dataset at every stage. Its totals are lower than those in \Cref{tab:data-pipeline-overview} because it covers only the 7 datasets retained for training, whereas \Cref{tab:data-pipeline-overview} counts all 10 downloaded datasets at the earlier stages.

\begin{table}[!t]
    \centering
    \scriptsize
    \begin{tabular}{lrrrr}
    \toprule
    \textbf{Dataset} & \textbf{Download} & \textbf{Render} & \textbf{Phase1 kept} & \textbf{Phase2 final} \\
    \midrule
    2wikimqa & 12,000 & 980 & 722 & 325 \\
    gov\_report & 12,000 & 8,000 & 8,000 & 966 \\
    hotpotqa & 12,000 & 2,650 & 2,235 & 1,270 \\
    lcc\_python & 9,995 & 8,000 & 4,604 & 1,035 \\
    musique & 12,000 & 8,000 & 5,704 & 2,080 \\
    qasper & 2,742 & 2,707 & 2,383 & 364 \\
    triviaqa & 20,000 & 8,000 & 6,830 & 4,001 \\
    \midrule
    \textbf{Total} & \textbf{80,737} & \textbf{38,337} & \textbf{30,478} & \textbf{10,041} \\
    \bottomrule
    \end{tabular}
    \caption{Per-dataset sample retention for the included training datasets from download to final filtering.}
    \label{tab:training-data}
\end{table}

The final training set in this paper uses 7 higher-volume datasets (\texttt{2wikimqa}, \texttt{gov\_report}, \texttt{hotpotqa}, \texttt{lcc\_python}, \texttt{musique}, \texttt{qasper}, \texttt{triviaqa}).
\texttt{dureader} and \texttt{qmsum} are excluded due to negligible retained Phase2 counts, and \texttt{narrativeqa} has no Phase2-final samples.

\paragraph{Text Rendering.}
We render text documents into images using ReportLab and pdf2image.
English text uses the Verdana font at 9pt font size, 72 DPI resolution, and A4 page size with 20px margins.
Each page is saved as a PNG image, and the corresponding source text is preserved for retrieval simulation during training.

\paragraph{Relevance Annotation.}
For each sample, we use Qwen3.5-9B to assess each rendered page independently and identify whether it is necessary for answering the query.
The model is prompted with the query, the ground-truth answer, and a single page's text, and asked to respond ``YES'' or ``NO'' for that page.
We use temperature=0.3 for consistency.
The ground-truth answer is used only during offline annotation, and is never provided during inference or evaluation.
Samples are kept in Phase1 only when \texttt{relevant\_indices} is non-empty.
Pages marked as relevant are used to construct the \texttt{<tool\_call>} output and \texttt{<|observation|>} content.

\paragraph{Reasoning Generation.}
Unlike approaches that provide the ground-truth answer to guide reasoning, we adopt an \emph{independent inference} strategy: the model generates both reasoning and answer without seeing the ground truth.
Given the query, textual context, and selected-image context, Qwen3-32B generates step-by-step reasoning within \texttt{<think>...</think>} tags, followed by the predicted answer.
We use temperature=0.7, top\_p=0.8, top\_k=20, and enable the model's native thinking mode for more detailed reasoning chains.

\paragraph{Quality Filtering.}
We apply a strict quality filter: \textbf{only samples where the model's predicted answer matches the ground-truth answer are retained}.
This increases the likelihood that retained reasoning chains are aligned with correct outcomes, rather than post-hoc rationalizations.
After Phase1 retention (30,697 samples), Phase2 filtering keeps 10,041 samples across 7 datasets used for training in this paper.
Notably, \texttt{narrativeqa} is not included in the Phase2-final pool.
This filtering improves trajectory quality, but it may also bias the training distribution toward samples that are easier for the teacher model.

\paragraph{Loss Mask Construction.}
For SFT training, we use the following dataset-level masking rule to compute loss only on model-generated tokens:
\begin{itemize}
    \item \texttt{<tool\_call>}...\texttt{</tool\_call>}: mask = 1 (page selection)
    \item \texttt{<|observation|>} and retrieved text: mask = 0 (system-provided)
    \item \texttt{<think>}...\texttt{</think>} and final answer: mask = 1 (reasoning and answer)
    \item All other tokens (user query, special tokens): mask = 0
\end{itemize}
This ensures the model learns to generate selections and reasoning, not to memorize system-provided observations.
On average, approximately 20--30\% of tokens in each sample have mask=1.

\subsection{Training Hyperparameters}
\label{app:hyperparameters}

\Cref{tab:hyperparameters} lists the hyperparameters used for training.
All experiments are conducted on 2$\times$ NVIDIA H100 80GB GPUs using the Verl framework with vLLM for inference.

\begin{table}[!t]
\centering
\small
\begin{tabular}{ll}
\toprule
\textbf{Hyperparameter} & \textbf{Value} \\
\midrule
Base model & Glyph-9B (full-parameter) \\
Learning rate & $3 \times 10^{-6}$ \\
Optimizer & AdamW, $\beta=(0.9, 0.95)$, weight decay $0.01$ \\
Warmup ratio & 0.05 \\
Gradient clipping & 1.0 \\
Batch size (total) & 2 \\
Micro batch size (per GPU) & 1 \\
Training steps & 6{,}000 ($\approx$1.2 epochs over 10{,}041 trajectories) \\
Max sequence length & 16,384 (right truncation) \\
Precision & bfloat16 \\
Strategy & FSDP (full shard), CPU offload disabled \\
Gradient checkpointing & True \\
Attention implementation & SDPA \\
Random seed & 42 \\
Tool-call loss weight & 2.0 \\
Format-tag loss weight & 1.2 \\
Answer-boundary loss weight & 1.0 \\
Reported checkpoint & final (step 6{,}000) \\
\bottomrule
\end{tabular}
\caption{Training hyperparameters. The three loss weights refer to the per-token weighting in
\cref{sec:training}: tokens inside the tool call and the structural format tags are upweighted
relative to ordinary generated tokens, while non-generated observation tokens are masked out
entirely.}
\label{tab:hyperparameters}
\end{table}

\subsection{Ablation Protocol: Without Retrieval (w/o Retrieve)}
\label{app:wo-retrieve}

For the ``w/o Retrieve'' ablation, the model performs selection but does not receive the retrieved text.
Instead, it must answer based solely on visual representations.
At inference time, after the model outputs \texttt{<tool\_call>}, we provide an observation that acknowledges the selection but does not include text content:

\begin{quote}
\small
\ttfamily
<|observation|{}>\\
You identified images [1, 3] as relevant. Please answer based on the visual content of these images.
\end{quote}

This ablation tests whether the value of our method comes from (1) the selection itself (knowing which images matter) or (2) the actual text retrieval (having precise textual content).
We use the same trained \textsc{Seer} model and only modify the observation at inference time, ensuring a fair comparison.
We refer to this condition as $k{=}0$ throughout the appendix.


\subsection{Remaining LongBench Task Results}
\label{app:longbench-subtasks}

\begin{table}[!htbp]
\centering
\resizebox{0.98\linewidth}{!}{%
\begin{tabular}{@{}l cccccccccc@{}}
\toprule
\multirow{2}{*}{\textbf{Model}}
& \multicolumn{2}{c}{\textbf{Single-Doc QA}}
& \multicolumn{2}{c}{\textbf{Multi-Doc QA}}
& \multicolumn{2}{c}{\textbf{Summarization}}
& \multicolumn{2}{c}{\textbf{Few-shot}}
& \multicolumn{2}{c}{\textbf{Code}} \\
\cmidrule(lr){2-3} \cmidrule(lr){4-5} \cmidrule(lr){6-7} \cmidrule(lr){8-9} \cmidrule(lr){10-11}
& \textbf{QA Zh} & \textbf{QA En} & \textbf{Mus} & \textbf{Dur} & \textbf{News} & \textbf{VCSUM} & \textbf{Sam} & \textbf{LSHT} & \textbf{RB} & \textbf{LCC} \\
\midrule
\multicolumn{11}{l}{\textit{\small Text-based Inference}} \\[-4pt]
\midrule
\textcolor{gray}{GPT-4.1} & \textcolor{gray}{63.90} & \textcolor{gray}{51.27} & \textcolor{gray}{55.63} & \textcolor{gray}{24.58} & \textcolor{gray}{23.70} & \textcolor{gray}{14.66} & \textcolor{gray}{41.25} & \textcolor{gray}{50.00} & \textcolor{gray}{67.94} & \textcolor{gray}{68.43} \\
\hdashline\addlinespace[2pt]
LLaMA-3.1-8B-Instruct & 62.20 & \textbf{54.98} & 31.61 & \textbf{33.75} & \underline{24.21} & \underline{16.23} & 7.61 & 0.00 & 42.81 & 46.35 \\
Qwen2.5-7B-Instruct-1M & \textbf{62.98} & 53.62 & 34.72 & 21.85 & 21.02 & 12.20 & \textbf{39.17} & 28.68 & 29.80 & 21.72 \\
Qwen3-8B & 60.64 & 47.61 & 47.32 & 16.89 & 18.45 & 12.19 & 37.06 & 41.83 & 40.71 & 41.84 \\
GLM-4-9B-Chat-1M & \underline{62.43} & 53.01 & 40.41 & \underline{27.07} & 23.71 & \textbf{16.52} & \underline{37.38} & \textbf{47.05} & 30.55 & 25.02 \\
\midrule
\multicolumn{11}{l}{\textit{\small Visual-Text Compression}} \\[-4pt]
\midrule
Glyph-9B & 37.82 & 43.18 & \underline{52.00} & 25.90 & 21.49 & 12.28 & 32.82 & \underline{45.50} & \underline{60.09} & \textbf{48.49} \\
\textbf{Seer (Ours)} & 56.99 & \underline{53.94} & \textbf{54.19} & 25.72 & \textbf{24.93} & 14.20 & 33.00 & 41.00 & \textbf{60.53} & \underline{48.31} \\
\bottomrule
\end{tabular}%
}
\caption{Remaining task-level LongBench results (\%); the main table is \Cref{tab:longbench-main}. Column abbreviations: QA~Zh/En $=$ MultiFieldQA-Zh/En, Mus $=$ MuSiQue, Dur $=$ DuReader, News $=$ MultiNews, Sam $=$ SAMSum, RB $=$ RepoBench-P. As in \Cref{tab:longbench-main}, the \textcolor{gray}{gray} GPT-4.1 row is a proprietary reference point excluded from the ranking; among the remaining models \textbf{bold} marks the best and \underline{underline} the second-best value per column.}
\label{tab:longbench-subtasks}
\end{table}

\subsection{Per-Task Token Breakdown}
\label{app:token-breakdown}

\Cref{tab:token-efficiency-breakdown-a} and \Cref{tab:token-efficiency-breakdown-b}
report the complete per-task token accounting for all 21 LongBench tasks under the same protocol used in the main text.
Each value is a per-sample mean for the corresponding task-model pair.

\begin{table}[t]
    \centering
    \scriptsize
    \setlength{\tabcolsep}{3.5pt}
    \resizebox{\textwidth}{!}{%
    \begin{tabular}{llrrrrrr}
    \toprule
    \textbf{Task} & \textbf{Model} & \multicolumn{4}{c}{\textbf{Prompt Tokens}} & \textbf{Completion} & \textbf{Compression ($\times$)} \\
    \cmidrule(lr){3-6}
    & & \textbf{Textual} & \textbf{Visual} & \textbf{Retrieved} & \textbf{Total} & & \\
    \midrule
    \texttt{2wikimqa} & GLM-4-9B-Chat-1M & 7197.465 & 0.000 & 0.000 & 7197.465 & 6.230 & 1.000 \\
    \texttt{2wikimqa} & Glyph-9B & 44.770 & 3106.955 & 0.000 & 3151.725 & 402.135 & 2.284 \\
    \texttt{2wikimqa} & Seer (Ours)$^{\dagger}$ & 2111.755 & 3331.335 & 1936.705 & 5443.090 & 478.390 & 1.322 \\
    \midrule
    \texttt{dureader} & GLM-4-9B-Chat-1M & 9803.400 & 0.000 & 0.000 & 9803.400 & 230.980 & 1.000 \\
    \texttt{dureader} & Glyph-9B & 22.695 & 1798.830 & 0.000 & 1821.525 & 201.465 & 5.382 \\
    \texttt{dureader} & Seer (Ours)$^{\dagger}$ & 4308.860 & 2021.125 & 4168.395 & 6329.985 & 550.070 & 1.549 \\
    \midrule
    \texttt{gov\_report} & GLM-4-9B-Chat-1M & 10302.390 & 0.000 & 0.000 & 10302.390 & 254.125 & 1.000 \\
    \texttt{gov\_report} & Glyph-9B & 40.000 & 3516.565 & 0.000 & 3556.565 & 929.930 & 2.897 \\
    \texttt{gov\_report} & Seer (Ours)$^{\dagger}$ & 5869.300 & 3741.615 & 5695.000 & 9610.915 & 2299.190 & 1.072 \\
    \midrule
    \texttt{hotpotqa} & GLM-4-9B-Chat-1M & 12898.870 & 0.000 & 0.000 & 12898.870 & 5.975 & 1.000 \\
    \texttt{hotpotqa} & Glyph-9B & 47.525 & 3755.995 & 0.000 & 3803.520 & 537.540 & 3.391 \\
    \texttt{hotpotqa} & Seer (Ours)$^{\dagger}$ & 3911.355 & 3981.415 & 3727.310 & 7892.770 & 400.890 & 1.634 \\
    \midrule
    \texttt{lcc} & GLM-4-9B-Chat-1M & 3195.778 & 0.000 & 0.000 & 3195.778 & 191.062 & 1.000 \\
    \texttt{lcc} & Glyph-9B & 21.000 & 2177.374 & 0.000 & 2198.374 & 224.276 & 1.454 \\
    \texttt{lcc} & Seer (Ours)$^{\dagger}$ & 969.236 & 2401.050 & 822.180 & 3370.286 & 634.740 & 0.948 \\
    \midrule
    \texttt{lsht} & GLM-4-9B-Chat-1M & 13319.445 & 0.000 & 0.000 & 13319.445 & 51.645 & 1.000 \\
    \texttt{lsht} & Glyph-9B & 515.985 & 4635.985 & 0.000 & 5151.970 & 356.270 & 2.585 \\
    \texttt{lsht} & Seer (Ours)$^{\dagger}$ & 3146.470 & 4863.030 & 2484.215 & 8009.500 & 386.680 & 1.663 \\
    \midrule
    \texttt{multi\_news} & GLM-4-9B-Chat-1M & 2647.955 & 0.000 & 0.000 & 2647.955 & 271.205 & 1.000 \\
    \texttt{multi\_news} & Glyph-9B & 36.000 & 829.785 & 0.000 & 865.785 & 1041.735 & 3.058 \\
    \texttt{multi\_news} & Seer (Ours)$^{\dagger}$ & 2902.185 & 1050.570 & 2757.475 & 3952.755 & 964.630 & 0.670 \\
    \midrule
    \texttt{multifieldqa\_en} & GLM-4-9B-Chat-1M & 6954.360 & 0.000 & 0.000 & 6954.360 & 24.113 & 1.000 \\
    \texttt{multifieldqa\_en} & Glyph-9B & 54.220 & 2636.540 & 0.000 & 2690.760 & 241.307 & 2.585 \\
    \texttt{multifieldqa\_en} & Seer (Ours)$^{\dagger}$ & 3220.453 & 2860.253 & 3039.953 & 6080.707 & 305.300 & 1.144 \\
    \midrule
    \texttt{multifieldqa\_zh} & GLM-4-9B-Chat-1M & 4148.700 & 0.000 & 0.000 & 4148.700 & 17.000 & 1.000 \\
    \texttt{multifieldqa\_zh} & Glyph-9B & 44.935 & 915.520 & 0.000 & 960.455 & 179.285 & 4.320 \\
    \texttt{multifieldqa\_zh} & Seer (Ours)$^{\dagger}$ & 3237.785 & 1136.395 & 3080.600 & 4374.180 & 340.065 & 0.948 \\
    \midrule
    \texttt{musique} & GLM-4-9B-Chat-1M & 15677.125 & 0.000 & 0.000 & 15677.125 & 6.145 & 1.000 \\
    \texttt{musique} & Glyph-9B & 46.600 & 6572.535 & 0.000 & 6619.135 & 730.000 & 2.368 \\
    \texttt{musique} & Seer (Ours)$^{\dagger}$ & 3345.990 & 6802.435 & 3135.990 & 10148.425 & 584.500 & 1.545 \\
    \midrule
    \texttt{narrativeqa} & GLM-4-9B-Chat-1M & 29871.560 & 0.000 & 0.000 & 29871.560 & 9.085 & 1.000 \\
    \texttt{narrativeqa} & Glyph-9B & 98.960 & 7775.110 & 0.000 & 7874.070 & 433.125 & 3.794 \\
    \texttt{narrativeqa} & Seer (Ours)$^{\dagger}$ & 6392.045 & 8006.845 & 6118.675 & 14398.890 & 366.985 & 2.075 \\
    \bottomrule
    \end{tabular}%
    }
    \caption{Per-task token breakdown on LongBench (Part I: 11 tasks). Values are per-sample means.}
    \label{tab:token-efficiency-breakdown-a}
\end{table}

\begin{table}[t]
    \centering
    \scriptsize
    \setlength{\tabcolsep}{3.5pt}
    \resizebox{\textwidth}{!}{%
    \begin{tabular}{llrrrrrr}
    \toprule
    \textbf{Task} & \textbf{Model} & \multicolumn{4}{c}{\textbf{Prompt Tokens}} & \textbf{Completion} & \textbf{Compression ($\times$)} \\
    \cmidrule(lr){3-6}
    & & \textbf{Textual} & \textbf{Visual} & \textbf{Retrieved} & \textbf{Total} & & \\
    \midrule
    \texttt{passage\_count} & GLM-4-9B-Chat-1M & 14908.325 & 0.000 & 0.000 & 14908.325 & 10.360 & 1.000 \\
    \texttt{passage\_count} & Glyph-9B & 93.000 & 4761.505 & 0.000 & 4854.505 & 1505.105 & 3.071 \\
    \texttt{passage\_count} & Seer (Ours)$^{\dagger}$ & 6680.260 & 4988.545 & 6441.020 & 11668.805 & 1186.850 & 1.278 \\
    \midrule
    \texttt{passage\_retrieval\_en} & GLM-4-9B-Chat-1M & 12546.610 & 0.000 & 0.000 & 12546.610 & 5.000 & 1.000 \\
    \texttt{passage\_retrieval\_en} & Glyph-9B & 239.850 & 3607.695 & 0.000 & 3847.545 & 153.040 & 3.261 \\
    \texttt{passage\_retrieval\_en} & Seer (Ours)$^{\dagger}$ & 2508.545 & 3832.865 & 2133.675 & 6341.410 & 272.490 & 1.979 \\
    \midrule
    \texttt{passage\_retrieval\_zh} & GLM-4-9B-Chat-1M & 4368.490 & 0.000 & 0.000 & 4368.490 & 4.020 & 1.000 \\
    \texttt{passage\_retrieval\_zh} & Glyph-9B & 130.095 & 1109.875 & 0.000 & 1239.970 & 142.035 & 3.523 \\
    \texttt{passage\_retrieval\_zh} & Seer (Ours)$^{\dagger}$ & 2769.525 & 1330.930 & 2528.100 & 4100.455 & 234.055 & 1.065 \\
    \midrule
    \texttt{qasper} & GLM-4-9B-Chat-1M & 5098.375 & 0.000 & 0.000 & 5098.375 & 16.340 & 1.000 \\
    \texttt{qasper} & Glyph-9B & 173.435 & 1580.635 & 0.000 & 1754.070 & 283.825 & 2.907 \\
    \texttt{qasper} & Seer (Ours)$^{\dagger}$ & 3955.465 & 1802.645 & 3665.970 & 5758.110 & 356.105 & 0.885 \\
    \midrule
    \texttt{qmsum} & GLM-4-9B-Chat-1M & 13610.870 & 0.000 & 0.000 & 13610.870 & 90.315 & 1.000 \\
    \texttt{qmsum} & Glyph-9B & 66.665 & 3818.290 & 0.000 & 3884.955 & 317.820 & 3.503 \\
    \texttt{qmsum} & Seer (Ours)$^{\dagger}$ & 5809.000 & 4043.830 & 5605.095 & 9852.830 & 402.020 & 1.381 \\
    \midrule
    \texttt{repobench-p} & GLM-4-9B-Chat-1M & 10842.382 & 0.000 & 0.000 & 10842.382 & 177.606 & 1.000 \\
    \texttt{repobench-p} & Glyph-9B & 1020.466 & 8227.592 & 0.000 & 9248.058 & 265.666 & 1.172 \\
    \texttt{repobench-p} & Seer (Ours)$^{\dagger}$ & 2399.458 & 8463.188 & 1181.416 & 10862.646 & 640.294 & 0.998 \\
    \midrule
    \texttt{samsum} & GLM-4-9B-Chat-1M & 9171.405 & 0.000 & 0.000 & 9171.405 & 44.115 & 1.000 \\
    \texttt{samsum} & Glyph-9B & 170.330 & 2295.920 & 0.000 & 2466.250 & 154.115 & 3.719 \\
    \texttt{samsum} & Seer (Ours)$^{\dagger}$ & 4152.455 & 2519.020 & 3859.525 & 6671.475 & 432.280 & 1.375 \\
    \midrule
    \texttt{trec} & GLM-4-9B-Chat-1M & 6795.650 & 0.000 & 0.000 & 6795.650 & 18.555 & 1.000 \\
    \texttt{trec} & Glyph-9B & 33.000 & 2114.625 & 0.000 & 2147.625 & 247.695 & 3.164 \\
    \texttt{trec} & Seer (Ours)$^{\dagger}$ & 5758.990 & 2337.460 & 5604.980 & 8096.450 & 349.225 & 0.839 \\
    \midrule
    \texttt{triviaqa} & GLM-4-9B-Chat-1M & 11819.240 & 0.000 & 0.000 & 11819.240 & 5.200 & 1.000 \\
    \texttt{triviaqa} & Glyph-9B & 747.520 & 3173.300 & 0.000 & 3920.820 & 186.150 & 3.014 \\
    \texttt{triviaqa} & Seer (Ours)$^{\dagger}$ & 4220.830 & 3397.800 & 3342.310 & 7618.630 & 282.675 & 1.551 \\
    \midrule
    \texttt{vcsum} & GLM-4-9B-Chat-1M & 8514.415 & 0.000 & 0.000 & 8514.415 & 248.970 & 1.000 \\
    \texttt{vcsum} & Glyph-9B & 23.000 & 1892.115 & 0.000 & 1915.115 & 443.645 & 4.446 \\
    \texttt{vcsum} & Seer (Ours)$^{\dagger}$ & 7403.125 & 2114.575 & 7248.365 & 9517.700 & 652.365 & 0.895 \\
    \bottomrule
    \end{tabular}%
    }
    \caption{Per-task token breakdown on LongBench (Part II: 10 tasks). Values are per-sample means.}
    \label{tab:token-efficiency-breakdown-b}
\end{table}


\subsection{Wall-Clock Efficiency and Long-Context Scaling}
\label{app:wallclock}

\Cref{sec:token-reduction} reports prompt-token compression, which is an architectural accounting
quantity rather than a measured runtime. Token counts are not a reliable proxy for latency, because
visual tokens are fewer but individually more expensive to produce. We therefore measure
prefill/time-to-first-token, end-to-end latency, decode throughput, and completion length directly.

\begin{table}[!htbp]
    \centering
    \small
    \setlength{\tabcolsep}{5pt}
    \resizebox{\textwidth}{!}{%
    \begin{tabular}{llccccccc}
    \toprule
    \multirow{2}{*}{\textbf{Model}} & \multirow{2}{*}{\textbf{Prompt form}}
    & \multicolumn{2}{c}{\textbf{Prefill (s)}} & \textbf{E2E}
    & \textbf{Decode} & \textbf{Completion} & \textbf{Prompt} & \multirow{2}{*}{\textbf{$n{>}30$s}} \\
    \cmidrule(lr){3-4}
    & & \textbf{server} & \textbf{client$+$server} & \textbf{median (s)}
    & \textbf{(tok/s)} & \textbf{tok (med.)} & \textbf{tok (mean)} & \\
    \midrule
    GLM-4-9B-Chat-1M & full text          & 0.23 & 0.23 & 1.07 & $\sim$76$^{\ddagger}$ & 8   & 5{,}811 & 1 \\
    Glyph-9B         & images             & 0.21 & 0.29 & 3.61 & 75.5 & 235 & 1{,}983 & 10 \\
    \textbf{\textsc{Seer}} & images $+$ retrieved text & 0.22 & 0.29 & 5.47 & 76.3 & 381 & 7{,}678 & 14 \\
    \bottomrule
    \end{tabular}%
    }
    \caption{Wall-clock measurements on LongBench under a controlled protocol: a single
    NVIDIA H100 80GB, batch size 1, 10 samples per task across all 21 tasks
    ($n=210/200/210$ completed samples), and the same checkpoints and Stage-2 pipeline as our main
    results. Prefill / time-to-first-token is measured with a cold \texttt{max\_tokens}$=$1 probe;
    \emph{server prefill} is the server-side time, while \emph{client$+$server prefill} additionally
    includes client-side image construction and transfer. Medians are reported because a small
    number of samples generate to the token cap; the count of samples exceeding 30\,s is given in the
    last column. $\ddagger$: GLM's raw decode rate (36 tok/s) is an artifact of its 8-token outputs;
    bucketed at a matched output length ($\sim$260 tokens) it also reaches $\sim$76 tok/s.}
    \label{tab:wallclock}
\end{table}

\begin{table}[!htbp]
    \centering
    \small
    \setlength{\tabcolsep}{8pt}
    \begin{tabular}{cc@{\hskip 2.5em}cc}
    \toprule
    \multicolumn{2}{c}{\textbf{Full text (GLM-4-9B-Chat-1M)}} & \multicolumn{2}{c}{\textbf{Rendered images (Glyph-9B)}} \\
    \cmidrule(lr){1-2}\cmidrule(lr){3-4}
    \textbf{Text tokens} & \textbf{Prefill (s)} & \textbf{Visual tokens} & \textbf{Prefill (s)} \\
    \midrule
    5.4k & 0.26 & 5.6k  & 0.78 \\
    20k  & 0.70 & 14.6k & 2.23 \\
    40k  & 1.89 & 29.4k & 4.75 \\
    79k  & 5.53 & 59k   & 11.5 \\
    --   & --   & 105k  & 26.0 \\
    \bottomrule
    \end{tabular}
    \caption{Prefill scaling with context length (batch size 1, cold measurements with distinct
    documents per repetition, so no prefix-cache reuse). Both text and visual prefill grow
    super-linearly. Because a document with $N$ text tokens renders to roughly $N/3$ visual tokens,
    the two columns should be compared at matched \emph{document} size: below roughly 90k text
    tokens (which covers all of LongBench), text prefill is comparable to or faster than the
    rendered form, so visual compression yields no single-request prefill speedup in this regime.
    Beyond that point the ordering reverses, and beyond roughly 128k text tokens the text form no
    longer fits at all while the rendered form still does. The visual column is measured through an
    external HTTP probe and therefore includes client-side image transfer (approximately 9\,MB at
    180 pages), so it is an upper bound on true server-side visual prefill and the real crossover
    is somewhat earlier.}
    \label{tab:prefill-scaling}
\end{table}

\Cref{tab:wallclock} reports the controlled benchmark. Three observations follow. First, prefill is
comparable across all three systems at LongBench context lengths (0.21--0.29\,s); there is no
prefill-latency win from visual compression in this regime. Second, decode throughput is identical
($\sim$76\,tok/s) because all three systems share the same 9B backbone, verified by bucketing the
full-text baseline by output length. Third, \textsc{Seer}'s higher end-to-end latency is explained
entirely by completion length: it emits an explicit reasoning trace (381 median tokens) where the
full-text baseline answers extractively (8 median tokens). We report medians because a small
fraction of samples generate to the token cap, and we give the count of samples exceeding 30\,s
rather than hiding that tail in a mean.

\Cref{tab:prefill-scaling} explains why the efficiency benefit is nonetheless real, but located
elsewhere. Text prefill grows super-linearly with context length (0.26\,s at 5.4k tokens to 5.53\,s
at 79k), and beyond roughly 128k tokens the text form does not fit at all while the rendered form
still does. Consistent with this, prior work reports its largest visual-compression speedups at
context lengths up to 128k and at maximum feasible batch size, which is a different operating point
from our batch-size-1 LongBench measurement; the two findings are compatible rather than
contradictory.

Accordingly, we state the efficiency claim as what the measurements support: visual compression
reduces the prompt and KV-cache footprint (approximately 67\% KV memory savings, which scale
linearly with tokens) and extends the context length that can be processed at all. It does not
deliver a single-request latency speedup at LongBench scale, and we do not claim one.

\subsection{Iso-Training Control: Glyph-9B-SFT-direct}
\label{app:iso-training}

\textsc{Seer} is fine-tuned on trajectories we construct, whereas the external text-based baselines
are evaluated in direct inference mode. A natural concern is that the reported gains come from
task-specific fine-tuning on our data rather than from the Select-Retrieve-Reason mechanism. To
separate the two, we train an iso-training control, \textbf{Glyph-9B-SFT-direct}, on exactly the
same trajectories, from the same Glyph-9B initialization, with the same optimizer, learning rate,
seed, batch size, maximum sequence length, loss, and 6{,}000-step budget. The single difference is
that the select, retrieve, and observation turns are stripped from every trajectory, so the model
learns to map rendered images directly to the answer. Its predictions are recovered with the same
three-round retry policy used for \textsc{Seer}, so the remaining failures are genuine degradations
rather than timeout artifacts.

\begin{table}[!htbp]
    \centering
    \small
    \setlength{\tabcolsep}{6pt}
    \begin{tabular}{lccc}
    \toprule
    \textbf{Model / condition} & \textbf{Select-Retrieve} & \textbf{LongBench Avg (\%)} & \textbf{QA-7 mean (\%)} \\
    \midrule
    Glyph-9B (off-the-shelf)                  & \ding{55} & 48.78 & 45.98 \\
    Glyph-9B-SFT-direct (iso-training control) & \ding{55} & 45.20 & 44.67 \\
    \textsc{Seer} without retrieval ($k{=}0$)  & selection only & 43.00 & -- \\
    \textbf{\textsc{Seer}}                     & \ding{51} & \textbf{51.11} & \textbf{52.54} \\
    \bottomrule
    \end{tabular}
    \caption{Iso-training control isolating the mechanism from the effect of fine-tuning on our
    data. \textbf{Glyph-9B-SFT-direct} is trained on exactly the same trajectories as
    \textsc{Seer}, from the same Glyph-9B initialization, with the same optimizer, learning rate,
    seed, batch size, maximum sequence length, loss, and step budget (6{,}000 steps); the only
    difference is that the select, retrieve, and observation turns are removed, so the model maps
    rendered images directly to an answer. \textbf{LongBench Avg} is the official category-level
    macro-average over all 21 tasks; \textbf{QA-7 mean} is the unweighted mean over the seven
    LongBench QA tasks (Qasper, NarrativeQA, MultiFieldQA-En/Zh, HotpotQA, 2WikiMQA, MuSiQue).
    Fine-tuning on our data alone does not reproduce \textsc{Seer}'s accuracy: both no-mechanism
    conditions land at 43--45, whereas the Select-Retrieve-Reason mechanism recovers $+5.91$ over
    the iso-training control and $+8.11$ over the $k{=}0$ condition.}
    \label{tab:iso-training}
\end{table}

\begin{table}[!htbp]
    \centering
    \small
    \setlength{\tabcolsep}{5pt}
    \begin{tabular}{lcccccc}
    \toprule
    \textbf{Model} & \textbf{Single-doc QA} & \textbf{Multi-doc QA} & \textbf{Summ.} & \textbf{Few-shot} & \textbf{Synthetic} & \textbf{Code} \\
    \midrule
    Glyph-9B            & 35.45 & 51.48 & 19.50 & 61.47 & 70.50 & 54.29 \\
    Glyph-9B-SFT-direct & 34.44 & 49.15 & 21.09 & 49.23 & 66.83 & 50.48 \\
    \textbf{\textsc{Seer}} & \textbf{46.18} & \textbf{52.20} & \textbf{21.67} & 60.44 & \textbf{71.75} & 54.42 \\
    \bottomrule
    \end{tabular}
    \caption{Category-level breakdown for the iso-training control. Direct fine-tuning without the
    select-retrieve turns over-specializes to the direct answer format and loses most of its
    accuracy on the few-shot category (LSHT $45.50 \to 22.25$, SAMSum $32.82 \to 9.60$), which is
    what drives its lower overall average. \textsc{Seer}, trained on the same data but with the
    mechanism retained, does not show this degradation, so the comparison is conservative with
    respect to our claim rather than favorable to it.}
    \label{tab:iso-training-categories}
\end{table}

\Cref{tab:iso-training} shows that fine-tuning on our data alone does not reproduce \textsc{Seer}'s
accuracy. Glyph-9B-SFT-direct reaches 45.20, below the 48.78 of off-the-shelf Glyph-9B, and the
independent no-mechanism condition ($k{=}0$) reaches 43.00; the mechanism recovers $+5.91$ and
$+8.11$ over these two conditions respectively. The same ordering holds on the seven QA tasks that
the method targets, where direct fine-tuning (44.67) does not improve on Glyph-9B (45.98) while
\textsc{Seer} reaches 52.54.

\Cref{tab:iso-training-categories} shows where the control loses its accuracy. Direct fine-tuning
over-specializes to the answer-only format and collapses on the few-shot category (LSHT
$45.50 \to 22.25$, SAMSum $32.82 \to 9.60$), which accounts for most of the gap to Glyph-9B.
\textsc{Seer}, trained on the same data with the mechanism retained, does not show this
degradation. We note this explicitly because it means the control is somewhat pessimistic as a
measure of ``what fine-tuning alone buys''. The conclusion we draw from it, namely that the
mechanism rather than the data produces the gain, is supported independently by the $k{=}0$
condition, which shares \textsc{Seer}'s weights and format exactly.

\subsection{Intrinsic Page-Selection Quality}
\label{app:selection-quality}

The main text evaluates selection only through downstream answer accuracy, which conflates the
quality of the selector with the quality of the answerer. To evaluate the selector on its own, we
annotate the LongBench evaluation set at page level and score each selection strategy directly.

\paragraph{Annotation protocol.}
We re-annotate all 4{,}750 evaluation examples (31{,}182 rendered pages) with the same
answer-conditioned protocol used for training-data construction: Qwen3.5-9B judges whether a page
is necessary to derive the answer, given the query, the gold answer, and the page text. These
labels are produced only for this analysis; they are never provided to \textsc{Seer} at training or
inference time, and re-annotating the evaluation split (rather than reusing training labels) avoids
any label leakage between the training and analysis pipelines. The annotation run completed with no
API errors and no unparsed pages. 3{,}007 of the 4{,}750 examples have at least one relevant page,
which is the subset on which page precision, recall, and F1 are well defined; we additionally report
F1 over all 4{,}750 examples, where an empty gold set is scored as requiring an empty selection.

\paragraph{Baselines.}
We compare \textsc{Seer}'s learned visual selector against random selection, first-page selection,
a BM25~\citep{robertson2009probabilistic} retriever over rendered page text, selecting all pages,
and oracle selection. Because page
count affects precision and recall directly, each baseline is given a per-example page budget
matched to the number of pages \textsc{Seer} selected for that example, with a one-page minimum;
this is why the match-$k$ baselines average 1.77 pages against \textsc{Seer}'s 1.62 on this subset
(1.59 over all 4{,}750 examples, as reported in \cref{sec:ablation-selection}).

\begin{table}[!htbp]
    \centering
    \small
    \setlength{\tabcolsep}{5pt}
    \begin{tabular}{lccccc}
    \toprule
    \textbf{Strategy} & \textbf{Page P} & \textbf{Page R} & \textbf{Page F1} & \textbf{All-sample F1} & \textbf{Selected pages} \\
    \midrule
    Random match-$k$ & 0.429 & 0.447 & 0.413 & 0.262 & 1.77 \\
    First-page match-$k$ & 0.507 & 0.518 & 0.486 & 0.308 & 1.77 \\
    All pages & 0.433 & 1.000 & 0.547 & 0.346 & 6.43 \\
    BM25 match-$k$ & \textbf{0.630} & \textbf{0.643} & \textbf{0.606} & 0.384 & 1.77 \\
    \textbf{\textsc{Seer}} & 0.592 & 0.620 & 0.578 & \textbf{0.425} & 1.62 \\
    \midrule
    \textit{Oracle gold pages} & 1.000 & 1.000 & 1.000 & 1.000 & 1.85 \\
    \bottomrule
    \end{tabular}

    \vspace{0.75em}
    \begin{tabular}{lccc}
    \toprule
    \textbf{Subset} & \textbf{\# examples} & \textbf{\textsc{Seer} F1} & \textbf{BM25 F1} \\
    \midrule
    Low query/evidence lexical overlap & 1,548 & 0.487 & 0.474 \\
    High query/evidence lexical overlap & 1,459 & 0.675 & 0.746 \\
    Summarization / abstractive tasks & 908 & 0.659 & 0.661 \\
    Distributed evidence ($>$3 gold pages) & 299 & 0.516 & 0.514 \\
    \bottomrule
    \end{tabular}
    \caption{Intrinsic page-selection analysis on the LongBench evaluation set. Gold page labels are newly annotated with the same answer-conditioned Qwen3.5-9B protocol used for training-data annotation, and are used only for analysis. The top block reports page precision, recall, and F1 on the 3,007 examples with at least one gold page, plus all-sample F1 over all 4,750 examples; the oracle row is the upper bound rather than a competing strategy. Each match-$k$ baseline is given the same per-example page budget as \textsc{Seer} (one-page minimum), which is why they average 1.77 pages against \textsc{Seer}'s 1.62. Bold marks the best non-oracle value per column. The bottom block compares \textsc{Seer} and BM25 on diagnostic subsets. BM25 is a strong lexical baseline and is slightly ahead on gold-nonempty page F1, while \textsc{Seer} selects fewer pages, is best on the all-sample metric, and is ahead of BM25 on low lexical-overlap examples.}
    \label{tab:selection-quality}
\end{table}

\paragraph{Findings.}
\Cref{tab:selection-quality} shows that the learned selector is substantive rather than incidental:
it is far above random match-$k$ ($0.578$ vs.\ $0.413$ F1) and first-page match-$k$ ($0.486$), so
the downstream gains cannot be explained by ``retrieving a couple of pages often enough.'' BM25 is a
strong lexical selector and is slightly higher on the gold-nonempty subset ($0.606$ vs.\ $0.578$);
we report this plainly rather than treating BM25 as a weak reference. The two selectors are
complementary along an interpretable axis: on examples with low lexical overlap between the query
and the evidence text, \textsc{Seer} is ahead ($0.487$ vs.\ $0.474$), while BM25 is clearly stronger
when the query terms appear verbatim in the evidence ($0.746$ vs.\ $0.675$). \textsc{Seer} is also
the best strategy on the all-sample metric ($0.425$), which additionally rewards abstaining on
examples with no page-localized evidence, and it achieves this while selecting the fewest pages of
any non-trivial strategy. On examples whose evidence is distributed over more than three gold pages,
both finite-budget selectors are recall-limited and essentially tied ($0.516$ vs.\ $0.514$),
which is the intrinsic counterpart of the distributed-evidence limitation stated in the Limitations
section.

\subsection{End-to-End Comparison with a BM25 Page Selector}
\label{app:bm25-downstream}

Intrinsic selection quality does not by itself establish that visual selection matters for the final
answer, since a selector can be marginally better at picking pages without changing what the
answerer produces. We therefore also compare the two selectors end to end, holding everything else
fixed: BM25 chooses the pages, and the same \textsc{Seer} Stage-2 answerer then generates the answer
from the same visual context, the same retrieval mechanism, and the same token budget. The systems
differ only in which pages are selected.

\begin{table}[!htbp]
    \centering
    \small
    \setlength{\tabcolsep}{6pt}
    \begin{tabular}{llcc}
    \toprule
    \textbf{Category} & \textbf{Task} & \textbf{BM25-forced selection} & \textbf{\textsc{Seer} selection} \\
    \midrule
    \multirow{4}{*}{Single-doc QA}
      & Qasper            & 46.23 & \textbf{47.74} \\
      & NarrativeQA       & 25.63 & \textbf{26.05} \\
      & MultiFieldQA-En   & 53.30 & \textbf{53.94} \\
      & MultiFieldQA-Zh   & 50.86 & \textbf{56.99} \\
    \midrule
    \multirow{4}{*}{Multi-doc QA}
      & HotpotQA          & \textbf{62.78} & 60.22 \\
      & 2WikiMQA          & 57.24 & \textbf{68.65} \\
      & MuSiQue           & 44.12 & \textbf{54.19} \\
      & DuReader          & \textbf{28.41} & 25.72 \\
    \midrule
    \multirow{2}{*}{Summarization}
      & GovReport         & 26.28 & \textbf{27.30} \\
      & VCSUM             & 2.21  & \textbf{14.20} \\
    \midrule
    \multirow{4}{*}{Few-shot}
      & TREC              & 78.00 & 78.00 \\
      & TriviaQA          & 84.47 & \textbf{89.76} \\
      & SAMSum            & \textbf{34.93} & 33.00 \\
      & LSHT              & 40.25 & \textbf{41.00} \\
    \midrule
    Code & LCC            & 25.71 & \textbf{48.31} \\
    \midrule
    \multicolumn{2}{l}{\textit{Mean over the 15 tasks above}}      & 44.03 & \textbf{48.34} \\
    \multicolumn{2}{l}{\textit{Mean excluding the code task (LCC)}} & 45.34 & \textbf{48.34} \\
    \bottomrule
    \end{tabular}
    \caption{End-to-end comparison of page selectors under an identical answering pipeline.
    Both columns use the same \textsc{Seer} Stage-2 answerer, the same retrieval mechanism, and
    the same token budget; the only difference is which pages are selected. In the left column
    the selection is forced to the pages returned by a BM25 retriever over rendered page text
    (with the per-example page budget matched to \textsc{Seer}); in the right column the pages
    are chosen by \textsc{Seer}'s learned visual selector. We report the 15 tasks for which the
    BM25-forced run completed under the same retry and merge policy as \textsc{Seer}; the
    remaining six tasks are excluded rather than reported from partial runs. BM25 is a strong
    lexical selector and wins on three lexically-driven tasks, but it degrades sharply where
    page relevance is not lexical (VCSUM) or where pages are code (LCC).}
    \label{tab:bm25-downstream}
\end{table}

\Cref{tab:bm25-downstream} reports the 15 tasks for which this run completed under the same retry
and merge policy used for \textsc{Seer}; the remaining six tasks are omitted rather than reported
from partial runs, so the comparison is restricted to matched, fully-completed tasks. \textsc{Seer}
selection yields 48.34 against 44.03 for BM25-forced selection over these 15 tasks (48.34 vs.\ 45.34
if the single code task is excluded). The per-task pattern matches the intrinsic analysis: BM25 is
competitive or better where relevance is lexical (HotpotQA, DuReader, SAMSum), while the learned
selector is substantially better on multi-hop QA (MuSiQue $+10.1$, 2WikiMQA $+11.4$), on
cross-lingual and layout-sensitive extraction (MultiFieldQA-Zh $+6.1$), and on tasks where
page relevance is not lexical at all (VCSUM, LCC). We read this as evidence that the two selectors
address different regimes, not that visual selection dominates lexical retrieval.

\subsection{Comparison with Recursive Language Models}
\label{app:rlm}

Recursive Language Models~\citep{zhang2025recursive} are a text-side approach to long-context
reasoning that, like
\textsc{Seer}, combines global navigation with local high-precision reading, but does so without
rendering or a vision-language model. We therefore compare against it directly.

\begin{table}[!htbp]
    \centering
    \small
    \setlength{\tabcolsep}{6pt}
    \begin{tabular}{lcccc}
    \toprule
    \textbf{Task} & \textbf{RLM-Qwen3-8B} & \textbf{Qwen3-8B (direct)} & \textbf{Glyph-9B} & \textbf{\textsc{Seer}} \\
    \midrule
    Qasper          & 35.61 & 43.86 & 38.99 & \textbf{47.74} \\
    NarrativeQA     & 16.65 & \textbf{27.08} & 21.83 & 26.05 \\
    MultiFieldQA-En & 43.58 & 47.61 & 43.18 & \textbf{53.94} \\
    MultiFieldQA-Zh & 50.72 & \textbf{60.64} & 37.82 & 56.99 \\
    HotpotQA        & 51.84 & \textbf{65.60} & 56.42 & 60.22 \\
    2WikiMQA        & 61.74 & \textbf{73.62} & 71.62 & 68.65 \\
    MuSiQue         & 38.04 & 47.32 & 52.00 & \textbf{54.19} \\
    \midrule
    \textbf{Mean}   & 42.60 & 52.25 & 45.98 & \textbf{52.54} \\
    \bottomrule
    \end{tabular}
    \caption{Comparison with Recursive Language Models on the seven LongBench QA tasks, where a
    text-side recursive reader is the applicable comparison. All systems are scored with the same
    official LongBench evaluator used elsewhere in this paper. RLM is run with the authors'
    released recursive framework and their post-trained checkpoint
    (\texttt{mit-oasys/rlm-qwen3-8b-v0.1}); the only code we add is a LongBench adapter that loads
    examples and writes predictions in LongBench format. Documents are not truncated: the full
    document is placed in RLM's local REPL as the context variable, and the root prompt keeps the
    LongBench QA template with the context field replaced by an instruction that the full context is
    available in the REPL. Decoding is greedy (temperature $0$, top-$p$ $1$) with
    \texttt{max\_tokens}$=$2048, \texttt{max\_depth}$=$1, and \texttt{max\_iterations}$=$8, served
    through a vLLM OpenAI-compatible endpoint; the checkpoint's native 40{,}960-token window applies
    per model call. The only RLM-specific post-processing is deterministic removal of wrapper labels
    such as ``FINAL ANSWER:'' before scoring, which raises rather than lowers RLM's score.}
    \label{tab:rlm}
\end{table}

\Cref{tab:rlm} reports the comparison on the seven LongBench QA tasks, where a text-side recursive
reader is the applicable comparison; code-completion and synthetic tasks are not. Under our scoring
protocol, RLM-Qwen3-8B reaches 42.60, below both direct Qwen3-8B (52.25) and \textsc{Seer} (52.54).

We do not read this as a negative result for RLM. Its reported strengths are on corpora spanning
hundreds of thousands to millions of tokens, where recursive decomposition is the only tractable
option and it improves substantially over its own base model. LongBench QA is a different regime:
contexts fit in a single window, and the tasks are dominated by targeted extraction and multi-hop
chains over one document, so decomposing the document into independent recursive sub-calls can
fragment exactly the cross-passage chains the task requires. \textsc{Seer} instead reads the whole
rendered document holistically and spends exact source text only on the few selected pages. The two
mechanisms operate on different axes (text-side recursion versus visual-text compression) and are
in principle combinable rather than mutually exclusive.

\subsection{Task-Wise Retrieval Heterogeneity Analysis}
\label{app:selection-bucket-alltasks}

\begin{figure}[t]
    \centering
    \includegraphics[width=0.98\linewidth]{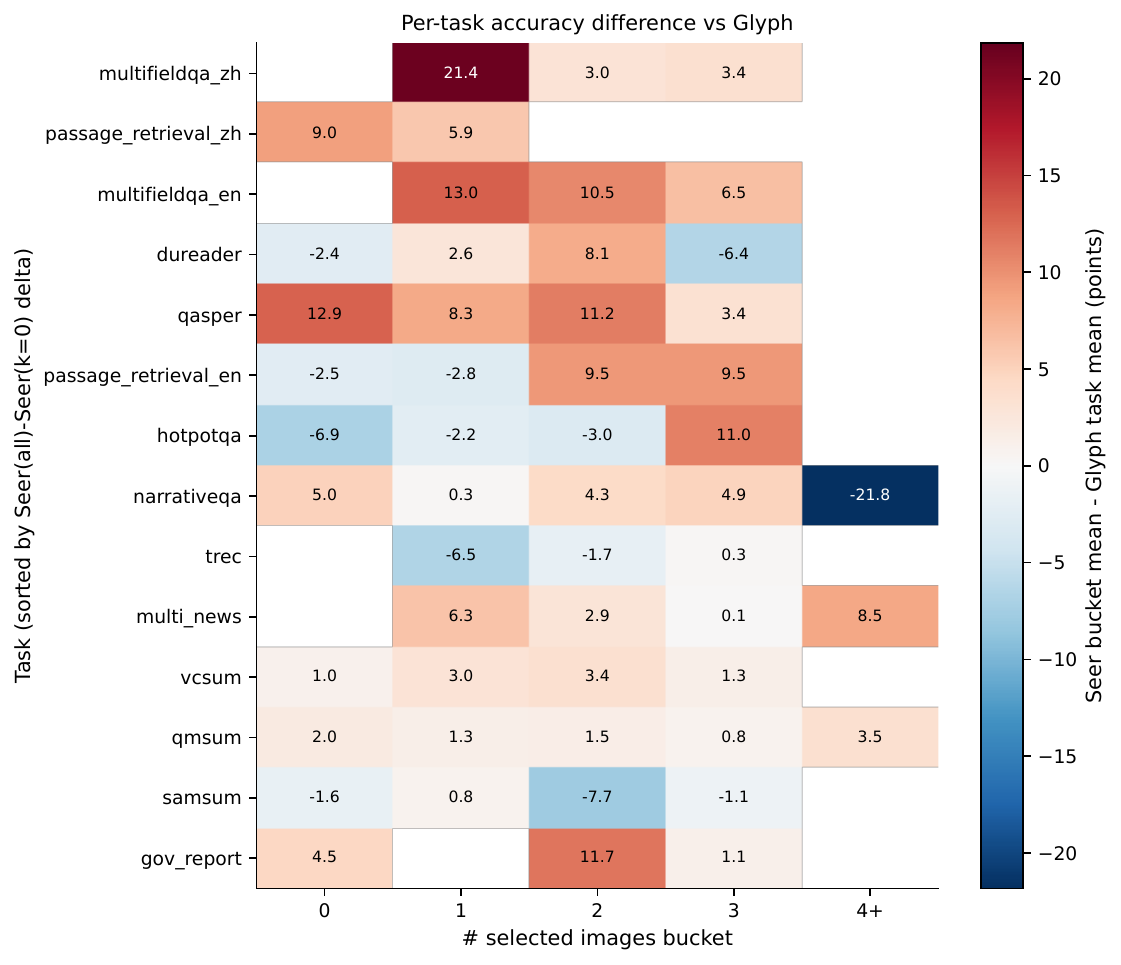}
    \caption{Task-wise retrieval heterogeneity view on LongBench, over the 14 tasks with sufficient per-bucket coverage. Color/value in each cell is \textit{Seer(full) bucket mean accuracy} minus \textit{Glyph task-level accuracy}; blank cells indicate empty buckets. Cells backed by few samples (e.g.\ the \texttt{narrativeqa} 4+ bucket) are correspondingly noisy and should not be read as task-level effects.}
    \label{fig:selection_bucket_alltasks_app}
\end{figure}

To complement the representative-task figure in the main text, \Cref{fig:selection_bucket_alltasks_app} reports the 14 LongBench tasks with sufficient per-bucket coverage for this view; the remaining tasks are omitted because their selected-image buckets are too sparsely populated to average meaningfully.
Each cell shows, for a given task and selected-image bucket, the difference between Seer(full) bucket-mean accuracy and the task-level Glyph score.
Tasks are sorted by the endpoint gain $\Delta = \text{Seer(all)} - \text{Seer}(k{=}0)$.
This view highlights that retrieval gains are highly task-dependent rather than uniform.

\subsection{Qualitative Success and Failure Cases}
\label{app:qualitative}

\newcolumntype{R}[1]{>{\raggedright\arraybackslash}p{#1}}
\begin{table}[!htbp]
    \centering
    \small
    \renewcommand{\arraystretch}{1.15}
    \setlength{\tabcolsep}{4pt}
    \begin{tabular}{@{}R{0.145\linewidth} R{0.29\linewidth} R{0.235\linewidth} R{0.235\linewidth}@{}}
    \toprule
    \textbf{Task} & \textbf{Query (abridged)} & \textbf{\textsc{Seer}} & \textbf{Glyph-9B} \\
    \midrule
    \multicolumn{4}{@{}l}{\textit{Successes: localized evidence, an exact list, magnitude, or polarity}} \\
    \addlinespace[2pt]
    Qasper &
    Which neural network modules are included in NeuronBlocks? &
    Embedding Layer, Neural Network Layers, Loss Function, Metrics \newline (matches gold) &
    Word/character embedding, RNN, CNN, Transformer, Focal Loss, Accuracy, Precision, F1, CRF, distillation \newline (invented list) \\
    \addlinespace[3pt]
    MultiFieldQA-Zh &
    Amount of local fiscal spending on social security over the past five years &
    650 million yuan \newline (matches gold) &
    over 5 billion yuan \newline (misread magnitude) \\
    \addlinespace[3pt]
    DuReader &
    Does a decayed wisdom tooth always have to be extracted? &
    Yes: wisdom teeth are hard to clean and hard to restore, so early extraction is advised \newline (matches gold) &
    No: it depends on severity and position \newline (contradicts gold) \\
    \midrule
    \multicolumn{4}{@{}l}{\textit{Failures: multi-hop questions whose evidence is distributed}} \\
    \addlinespace[2pt]
    MuSiQue &
    Multi-hop bridge question; gold is \emph{Manhattan Project} &
    \emph{7th Solvay Conference} \newline (correct intermediate hop, chain not completed) &
    -- \\
    \addlinespace[3pt]
    MuSiQue &
    Multi-hop role chain; gold is \emph{Maria Bello} &
    \emph{Salma Hayek} \newline (entity resolved to the wrong person) &
    -- \\
    \bottomrule
    \end{tabular}
    \caption{Representative successes and failures, taken verbatim from the \textsc{Seer} and
    Glyph-9B prediction files. The successes are the intended regime: the answer hinges on an exact
    list, number, or polarity that is present on one or two pages, and retrieving the source text
    for those pages resolves what purely visual decoding gets wrong. The failures are the regime our
    analysis identifies as hardest: multi-hop questions whose evidence is distributed, where a single
    round of selection can surface a correct intermediate hop without completing the chain. A dash
    in the last column marks a case we report only for \textsc{Seer}. This is the qualitative
    counterpart to the distributed-evidence subset in \Cref{tab:selection-quality}, and it motivates
    iterative selection as future work.}
    \label{tab:qualitative}
\end{table}

\Cref{tab:qualitative} gives representative cases drawn verbatim from the prediction files. The
successes are the regime the method targets: the answer depends on an exact list, a numeric
magnitude, or a polarity that appears on one or two pages, and retrieving the source text for those
pages fixes what purely visual decoding misreads. In the Qasper case, \textsc{Seer} reproduces the
gold module list while Glyph-9B invents a plausible but incorrect one; in the MultiFieldQA-Zh case
the two answers differ by an order of magnitude, a characteristic failure of reading densely
rendered Chinese text visually.

The failures are equally informative and concentrate on multi-hop questions with distributed
evidence. In one MuSiQue example \textsc{Seer} retrieves and correctly answers an intermediate hop
but does not chain to the final entity; in another it resolves a multi-hop role chain to the wrong
person. A single round of selection is well suited to localized evidence and under-serves questions
whose evidence must be assembled across many pages, which is the same conclusion the
distributed-evidence subset in \Cref{tab:selection-quality} reaches quantitatively.

\end{document}